\documentclass[a4paper]{styles/svproc}
\usepackage{url}
\usepackage{cite}

\usepackage{graphics} % for pdf, bitmapped graphics files
\usepackage{epsfig} % for postscript graphics files
\usepackage{mathptmx} % assumes new font selection scheme installed
\usepackage{times} % assumes new font selection scheme installed
\usepackage{amsmath} % assumes amsmath package installed
\usepackage{amssymb}  % assumes amsmath package installed
\usepackage{todonotes}

\usepackage{appendix}

\usepackage{subcaption}

\usepackage{hyperref}

\usepackage{sidecap}

\usepackage[font=small,labelfont=bf]{caption}
\usepackage[resetlabels,labeled]{multibib}
\newcites{Appendix}{References for Appendix}

\usepackage{etoolbox}
\newtoggle{appendix}
\toggletrue{appendix}

\title{\LARGE \bf
kPAM: KeyPoint Affordances for Category-Level Robotic Manipulation
} 

\author{Lucas Manuelli*, Wei Gao*, Peter Florence, Russ Tedrake}
\institute{ 
CSAIL, Massachusetts Institute of Technology,
\\ 
\{manuelli, weigao, peteflo, russt\}@mit.edu
\\
\textit{*These authors contributed equally to this work.}}

\begin{document}

\maketitle
\thispagestyle{empty}
\pagestyle{plain}

%%%%%%%%%%%%%%%%%%%%%%%%%%%%%%%%%%%%%%%%%%%%%%%%%%%%%%%%%%%%%%%%%%%%%%%%%%%%%%%%
\begin{abstract}

We would like robots to achieve purposeful manipulation by placing any instance from a category of objects into a desired set of goal states.
Existing manipulation pipelines typically specify the desired configuration as a target 6-DOF pose and rely on explicitly estimating the pose of the manipulated objects.
However, representing an object with a parameterized transformation defined on a fixed template cannot capture large intra-category shape variation, and specifying a target pose at a category level can be physically infeasible or fail to accomplish the task -- e.g. knowing the pose and size of a coffee mug relative to some canonical mug is not sufficient to successfully hang it on a rack by its handle.
Hence we propose a novel formulation of category-level manipulation that uses semantic 3D keypoints as the object representation. This keypoint representation enables a simple and interpretable specification of the manipulation target as geometric costs and constraints on the keypoints, which flexibly generalizes existing pose-based manipulation methods. Using this formulation, we factor the manipulation policy into instance segmentation, 3D keypoint detection, optimization-based robot action planning and local dense-geometry-based action execution. This factorization allows us to leverage advances in these sub-problems and combine them into a general and effective perception-to-action manipulation pipeline. Our pipeline is robust to large intra-category shape variation and topology changes as the keypoint representation ignores task-irrelevant geometric details.
Extensive hardware experiments demonstrate our method can reliably accomplish tasks with never-before seen objects in a category, such as placing shoes and mugs with significant shape variation into category level target configurations. The video, supplementary material and source code are available on our project page \href{https://sites.google.com/view/kpam}{\textcolor{blue}{\underline{https://sites.google.com/view/kpam}}}.

\end{abstract}

%%%%%%%%%%%%%%%%%%%%%%%%%%%%%%%%%%%%%%%%%%%%%%%%%%%%%%%%%%%%%%%%%%%%%%%%%%%%%%%%
\section{Introduction}

This paper focuses on pose-aware robotic pick and place at a category level. Contrary to single-instance pick and place, the manipulation policy should generalize to potentially unknown instances in the category with different shape, size, appearance, and topology. These tasks can be easily described using natural language, for example ``put the mugs upright on the shelf,'' ``hang the mugs on the rack by their handle'' or ``place the shoes onto the shoe rack.'' However, converting these intuitive descriptions into concrete robot actions remains a significant challenge. Accomplishing these types of tasks is of significant importance to both industrial applications and interactive assistant robots.

While a large body of work addresses robotic picking for arbitrary objects~\cite{gualtieri2016gpd, zeng2018affordance, pmlr-v87-kalashnikov18a}, existing methods have not demonstrated pick \textit{and} place with an interpretable and generalizable approach.  One way to achieve generalization at the object category level, and perhaps the most straightforward approach is to attempt to extend existing instance-level pick and place pipelines with category-level pose estimators %for example those that aim to produce a category level pose by training on data annotated with pre-aligned geometric templates 
~\cite{sahin2018category, wang2019normalized}. %. With advances in deep learning, carefully designed neural networks~\cite{sahin2018category, tremblay2018deep, wang2019normalized} can produce accurate and robust pose estimation. %Several works~\cite{sahin2018category, wang2019normalized} aim to produce a category level pose by training on data annotated with pre-aligned geometric templates. 
However, as detailed in Sec.~\ref{sec:comparison}, representing an object with a parameterized pose defined on a fixed geometric template, as these works do, may not adequately capture large intra-class shape or topology variations, and can lead to physically infeasible target pose for certain instances in the category.  Other recent work has developed dense correspondence visual models, including at a category level, as a general representation for robot manipulation \cite{florence2018dense}, but did not formulate how to specify and solve the task of manipulating objects into specific configurations.  As a different route to address category-level pick and place, without an explicit object representation, \cite{gualtieri2018pick} trains end-to-end policies in simulation to generalize across the object category. It is unclear, however, how to measure the reward function for this type of approach in a fully general way without an object representation that can adequately capture the human's intention for the task.

%which may well be suited to an interpretable object representation.

%defining a target pose for category level robotic pick and place tasks can be ambiguous and/or lead to poses that are physically physically infeasible for certain instances of a category. In addition, pose estimation approaches may not adequately capture large intra-class shape or topology variations. 
%As a result, extending existing pose-based pick and place pipelines with category level pose estimators does not directly solve the task.
%
%
%Another line of work, usually referred to as robotic grasping algorithms, aims to solve the problem of finding stable grasps that can reliably pick up the objects~\cite{gualtieri2016gpd, morrison2018closing, zeng2018affordance}. Many of these approaches~\cite{gualtieri2016gpd, zeng2018affordance} are purely geometric and agnostic to the object class. When combined with instance segmentation~\cite{he2017maskrcnn}, these methods can accomplish tasks such as ``pick up a specific object". 
%Although these methods have demonstrated promising generality, due to the fact that they have no notion of the pose of an object, they struggle to accomplish any tasks beyond simple grasping of a specific object and placing it arbitrarily. Hence many important applications which require the manipulated objects to be placed in specific goal states are out of scope for these methods.
%

\begin{figure}[t]
\centering
\includegraphics[width=0.7\textwidth]{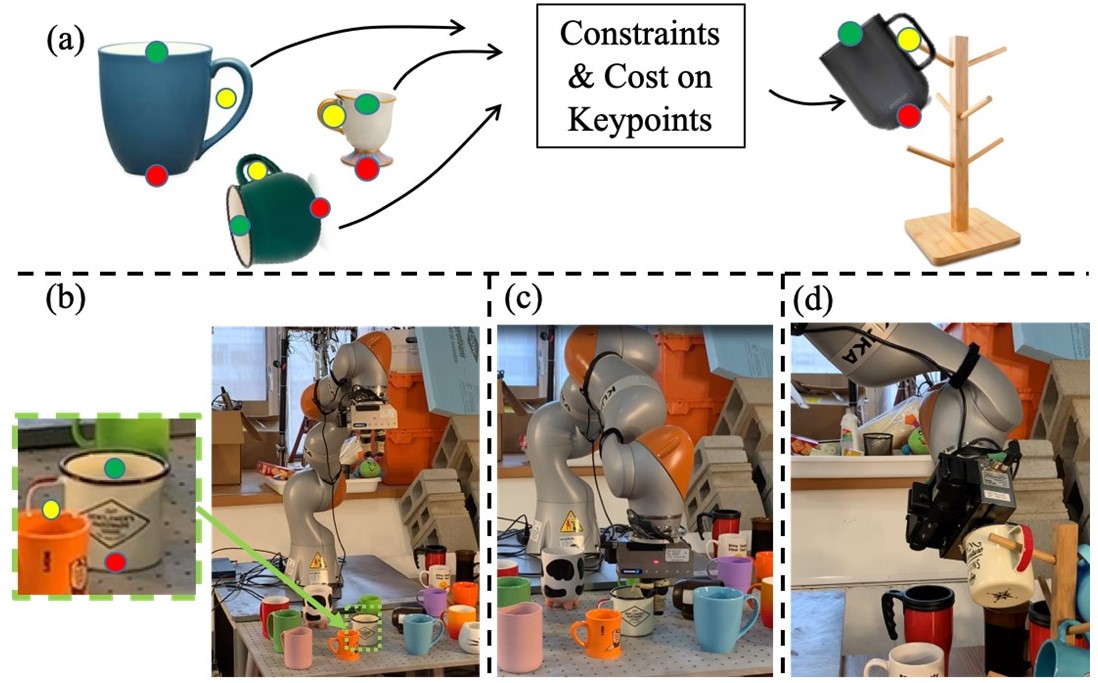}.
\caption{\label{fig:kPAM_fig_1} 
kPAM is a framework for defining and accomplishing category level manipulation tasks. The key distinction of kPAM is the use of semantic 3D keypoints as the object representation (a), which enables flexible specification of manipulation targets as geometric costs/constraints on keypoints. Using this framework we can handle wide intra-class shape variation (a) and reliably accomplish category-level manipulation tasks such as perceiving (b), grasping (c), and (d) placing any mug on a rack by its handle. A video demo for this task is available on our \href{https://sites.google.com/view/kpam}{\textcolor{blue}{\underline{project page}}}.
}
\end{figure}

\textbf{Contributions.} Our main contribution is a novel formulation of the category-level pick and place task which uses semantic 3D keypoints as the object representation. 
This keypoint representation enables a simple and interpretable specification of the manipulation target as geometric costs and constraints on the keypoints, which flexibly generalizes existing pose-based manipulation targets.
Using this formulation, we contribute a manipulation pipeline that factors the problem into 1) instance segmentation, 2) 3D keypoint detection, 3) optimization-based robot action planning 4) geometric grasping and action execution. This factorization allows us to leverage well-established solutions for these submodules and combine them into a general and effective manipulation pipeline. 
The keypoint representation ignores task-irrelevant geometric details of the object, making our method robust to large intra-category shape and topology variations.
% In addition to the usage in our pipeline, the keypoint representation can potentially contribute to various learning-based manipulation approaches as 1) a reward function to flexibly specify the manipulation target or 2) an alternative input to the policy/value neural network, which is more robust to shape variation and large deformation than the widely-used pose representation.
We experimentally demonstrate the use of this keypoint representation with our manipulation pipeline on several category-level pick and place tasks implemented on real hardware. We show that our approach generalizes to novel objects in the category, and that this generalization is accurate enough to accomplish tasks requiring centimeter level precision. 

\textbf{Paper Organization.} In Sec.~\ref{sec:related_works} we review related work. Sec.~\ref{sec:formulation} describes our formulation of category-level manipulation tasks. The formulation is introduced in Sec.~\ref{subsec:formulation_example} using a concrete example, while Sec.~\ref{subsec:general_formulation} describes the general formulation. Sec.~\ref{sec:comparison} compares our formulation with pose-based pick and place pipelines, highlighting the flexibility and generality of our method. Sec.~\ref{sec:results} describes the results of hardware experiments on 3 different category-level manipulation tasks, specifically showing generalization to novel object instances. Sec.~\ref{sec:limitations} discusses limitations and future work and Sec.~\ref{sec:conclusion} concludes.

% \vspace{-1em}
\section{Related Work}
\label{sec:related_works}

\subsection{Object Representations and Perception for Manipulation}

There exist a number of object representations, and methods for perceiving these representations, that have been demonstrated to be useful for robot manipulation. For a pick and place task involving a known object the standard solution starts by estimating the object's 6DOF pose. This allows the robot to then move the object from it's estimated pose to the specified target pose. Pose estimation is an extensively studied topic in computer vision and robotics, and existing methods can be generally classified into geometry-based algorithms~\cite{myronenko2010cpd, gao2019filterreg} and learning-based approaches~\cite{tremblay2018deep, wang2019normalized, sahin2018category}. There exist several datasets ~\cite{wang2019normalized, xiang2014beyond} annotated with aligned geometric templates, and pose estimators~\cite{sahin2018category, wang2019normalized} trained on these datasets can produce a category-level pose estimation.  Consequently, a straightforward approach to category-level pose-aware manipulation is to combine single object pick and place pipelines with these perception systems. However, pose estimation can be ambiguous under large intra-category shape variations, and moving the object to the specified target pose for the geometric template can lead to incorrect or physically infeasible states for different instances within a category of objects. For example knowing the pose and size of a coffee mug relative to some canonical mug is not sufficient to successfully hang it on a rack by its handle. A more technical discussion is presented in Sec.~\ref{sec:comparison}.

Other work has developed and used representations that may be more generalizable than object-specific pose estimation.  Recent work has demonstrated dense visual descriptors \cite{schmidt2017} as a fully self-supervised object representation for manipulation that can generalize at the category level \cite{florence2018dense}. In comparison with our present work based on 3D keypoints: (i) it is unclear how to extend dense visual descriptors to represent the full object configuration due to self-occlusions which would require $N$ layers of occluded descriptors, (ii) the sparse keypoint representation may in practice be more effective at establishing task-relevant correspondence across significant topology variation, and (iii) correspondence alone may not fully define a class-general configuration-change manipulation task, but the addition of human-specified geometric costs and constraints on 3D keypoints may. Keypoints have also been used in prior works as components of manipulation pipelines. Several prior works demonstrate the manipulation of deformable objects, and keypoint detection plays a role in their respective perception pipelines. The detected keypoints are typically used as grasp points~\cite{maitin2010cloth, seita2018bed} or building blocks for other shape parameterizations, e.g. the polygons in~\cite{van2010gravity, miller2011parametrized, miller2012geometric} on which the manipulation policy is defined. These approaches tackled various challenging manipulation tasks such as bed making and towel folding. In contrast, we propose a novel category-level manipulation target specification using costs and constraints defined on 3D keypoints. Additionally the state-machine approach in \cite{maitin2010cloth, seita2018bed} and manipulation primitives of \cite{van2010gravity, miller2012geometric} are specific to cloth and hence our manipulation task is out of scope for these approaches.

% \vspace{-1em}
\subsection{Grasping Algorithms}

% Grasping algorithms enable finding stable grasp poses that allow robots to reliably pick up objects.
In recent years there have been significant advances in grasping algorithms that allow robots to reliably pick up a wide range of objects, including potentially unseen objects. Among various approaches for grasping, model-based methods~\cite{zeng2017multi, mahler2016dex} typically rely on a pre-built grasp database of common 3D object models labeled with sets of feasible grasps. During execution, these methods associate the sensor input with an object entry in the database for grasp planning. In contrast, model-free methods~\cite{zeng2018affordance, gualtieri2016gpd, mahler2019learning} directly evaluate the grasp quality from raw sensor inputs. Many of these approaches achieved promising robustness and generality in the Amazon Picking Challenge~\cite{zeng2017multi, schwarz2018fast, zeng2018affordance}. Several works also incorporate object semantic information using instance masks~\cite{schwarz2018fast}, or non-rigid registrations~\cite{rodriguez2018transferring} to accomplish tasks such as picking up a specific object  or transferring a grasp pose to novel instances.  

In this work we focus on category-level manipulation tasks which require placing novel instances of a category into desired goal states. Although the ability to reliably grasp an object is an important part of our manipulation pipeline, it doesn't help with the problem of deciding what to do with the object after it has been grasped. Thus the tasks that we consider are out of scope for the aforementioned grasping works.

% \vspace{-1em}
\subsection{End-to-End Reinforcement Learning}

There have been impressive contributions~\cite{gualtieri2018pick, andrychowicz2018learning} in end-to-end reinforcement learning with applications to robotic manipulation. In particular, \cite{gualtieri2018pick} has demonstrated robotic pick and place across different instances and is the most related to our work. These end-to-end methods encode a manipulation task into a reward function and train the policy using trial-and-error.

However, in order to accomplish the category level pose-aware manipulation task, these end-to-end methods lack a general, flexible, and interpretable way to specify the desired configuration, which is required for the reward function. In~\cite{gualtieri2018pick}, the target configuration is implemented specific to the demonstrated task and object category. Extending it to other desired configurations, object categories and tasks is not obvious.  In this way, using end-to-end reinforcement learning allows the policy to be learned from experience without worrying about the details of shape variation, but only transfers the burden of shape variation to the choice and implementation of the reward function. We believe that our proposed object representation of 3D keypoints could be used as a solution to this problem.   %Furthermore, changing the desired configurations (in other words the reward function) requires retraining the policy network, which can be resource intensive even in simulation. \pete{This is not true since the policy can be goal-conditioned and finally there is work that uses goal-conditioned policies: https://learning-from-play.github.io/}
\begin{figure}[t]
\centering
\includegraphics[width=0.8\textwidth]{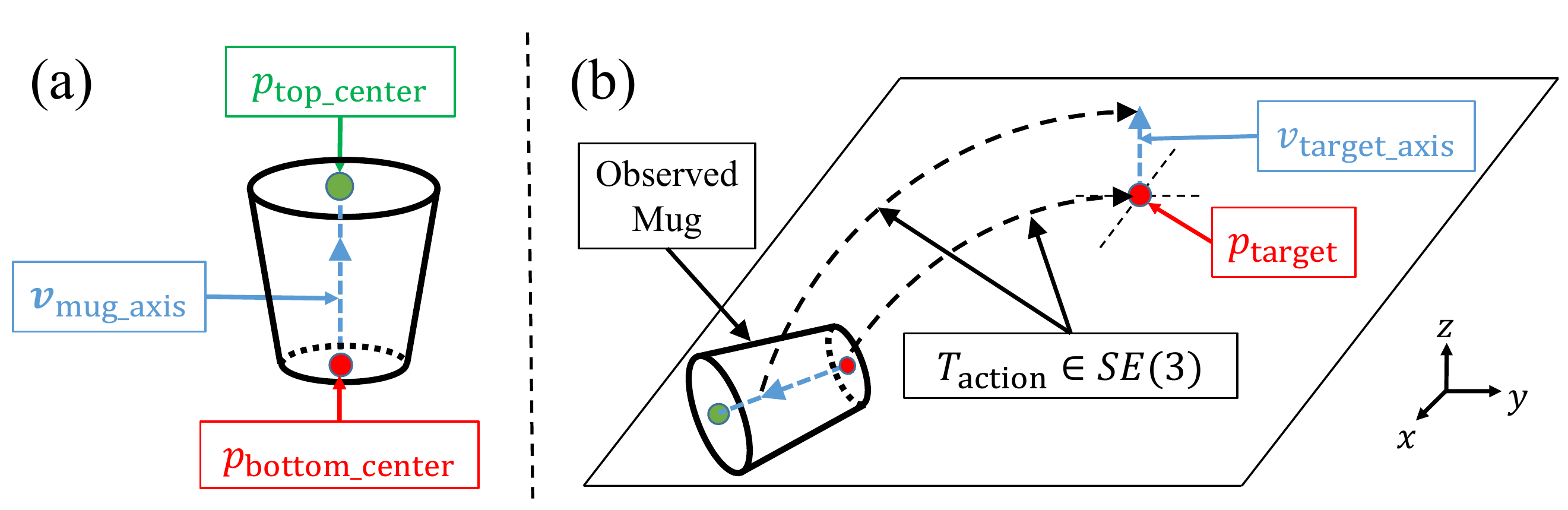}
\caption{\label{fig:formulation} An overview of our manipulation formulation using the ``put mugs upright on the table'' task as an example: 
(a) we train a category level keypoint detector that produces two keypoints: $p_\text{bottom\_center}$ and $p_\text{top\_center}$. The axis of the mug $v_\text{mug\_axis}$ is a unit vector from $p_\text{bottom\_center}$ to $p_\text{top\_center}$. 
(b) Given an observed mug, its two keypoints on bottom center and top center are detected. The rigid transform $T_\text{action}$, which represents the robotic pick-and-place action, is solved to move the bottom center of the mug to the target location $p_\text{target}$ and align the mug axis with the target direction $v_\text{target\_axis}$. }
\end{figure}

\section{Manipulation Formulation}
\label{sec:formulation}
In this section, we describe our formulation of the category level manipulation problem. Sec. ~\ref{subsec:formulation_example} describes the approach using a concrete example while Sec. ~\ref{subsec:general_formulation} presents the general formulation.

% \vspace{-1em}
\subsection{Concrete Motivating Example}
\label{subsec:formulation_example}

Consider the task of ``put the mug upright on the table". We want to come up with a manipulation policy that will accomplish this task for mugs with different size, shape, texture and topology. 

To accomplish this task, we pick 2 semantic keypoints on the mugs: the bottom center $p_\text{bottom\_center}$ and the top center $p_\text{top\_center}$, as shown in Fig.~\ref{fig:formulation} (a). Additionally, we assume we have a keypoint detector, discussed in Section \ref{subsec:general_formulation}, that takes as input raw observations (typically RGBD images or point clouds) and outputs the 3D locations of the specified keypoints. Note that there is no restriction that the keypoints be on the object surface, as evidenced by keypoint $p_\text{top\_center}$ in Fig.~\ref{fig:formulation} (a). The 3D keypoints are usually expressed in the camera frame, but they can be transformed to an arbitrary frame using the known camera extrinsics. In the following text, we use $\mathbf{p} = \{{p_i}\}_{i=1}^N \in R^{3 \times N}$ to denote the detected keypoint positions in world frame, where $p_i$ is the $i^{\text{th}}$ detected keypoint, and $N$ is the total number of keypoints. In this example $N=2$.

For robotic pick-and-place of mostly rigid objects, we represent the robot action as a rigid transform $T_\text{action}$ on the manipulated object. Thus, the keypoints associated with the manipulated object will be transformed as $T_\text{action} \mathbf{p} \in R^{3 \times N}$ using the robot action. In practice, this action $T_\text{action}$ is implemented by first grasping the object using the algorithm detailed in Sec.~\ref{subsec:general_formulation} and then planning and executing a trajectory which ends with the object in the desired target location. This trajectory may require approaching the target from a specific direction, for example in the ``mug upright on the table'' task the mug must approach the table from above.

Given the above analysis, the manipulation task we want to accomplish can be formulated as finding a rigid transformation $T_\text{action}$ such that
\begin{enumerate}
\item The transformed mug bottom center keypoint should be placed at some target location: 
\begin{equation}
\label{equ:mug_target_position}
||T_\text{action} p_\text{bottom\_center} - p_\text{target}|| = 0
\end{equation}
\item The transformed direction from the mug bottom center to the top center should be aligned with the upright direction. This is encoded by adding a cost to the objective function
\begin{equation}
\label{equ:mug_target_axis}
 ||1 - \langle v_\text{target\_axis}, \hspace{0.05in} \text{rot}(T_\text{action})v_\text{mug\_axis} 
 \rangle||^2
\end{equation}
\noindent where $\text{rot}(T)$ is the rotational component of the rigid transformation $T$, the target orientation $v_\text{target\_axis} = [0, 0, 1]^{T}$, and
\begin{equation} 
 v_\text{mug\_axis} = \frac{p_\text{top\_center} - p_\text{bottom\_center}}{||p_\text{top\_center} - p_\text{bottom\_center}||}
\end{equation}
\end{enumerate}
\noindent An illustration is presented in Fig.~\ref{fig:formulation} (b). The above problem is an inverse kinematics problem with $T_\text{action}$ as the decision variable, a constraint given by Eq.~(\ref{equ:mug_target_position}) and cost given by Eq.~(\ref{equ:mug_target_axis}). This inverse kinematics problem can be reliably solved using off-the-shelf optimization solvers such as~\cite{drake}. We then pick up the object using robotic grasping algorithms~\cite{mahler2019learning, gualtieri2016gpd} and execute a robot trajectory which applies the manipulation action $T_\text{action}$ to the grasped object.
\begin{figure}[t]
\centering
\includegraphics[width=0.9\textwidth]{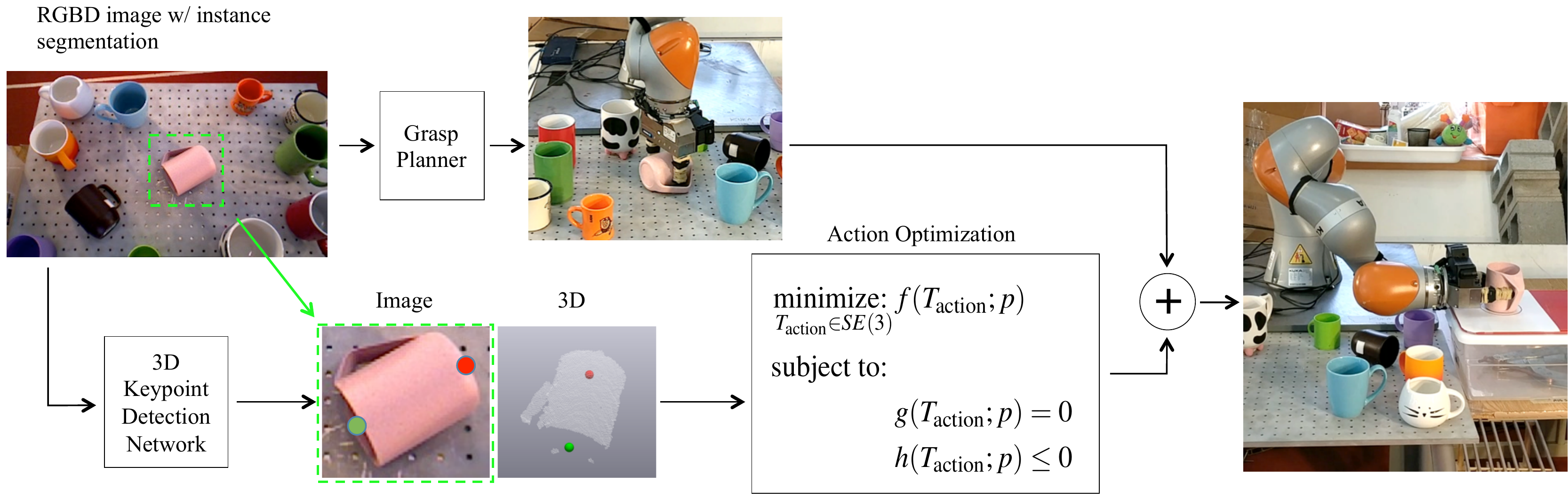}
\caption{\label{fig:pipeline_figure} 
An overview of the category level pick and place pipeline using our manipulation formulation. Given a RGBD image with instance segmentation, the semantic 3D keypoints of the object in question are detected. We then feed these 3D keypoints into an optimization based planning algorithm to compute the robot pick and place actions, which is represented by a rigid transformation $T_\text{action}$. Finally, we use an object-agnostic grasp planner to pick up the object and apply the computed robot action.
}
\end{figure}

\subsection{General Formulation}
\label{subsec:general_formulation}

Given an arbitrary category-level manipulation task we propose to solve it in the following manner. First the modeler must specify a set of semantic 3D keypoints $\mathbf{p} = \{{p_i}\}_{i=1}^N$ for the category, together with a set of geometric costs and constraints on these keypoints which fully specify the task, e.g. Eq. (\ref{equ:mug_target_position})-(\ref{equ:mug_target_axis}). It is up to the modeler to choose the keypoints, costs and constraints that encode the task. This step can be seen as analogous to choosing costs and constraints in a trajectory optimization or planning method, or specifying a reward function in a reinforcement learning approach. The only restriction on the choice of the 3D keypoints is that they must be well defined for all instances of the category that might be encountered at test time. In particular the 3D keypoints need not lie on the object surface, as demonstrated by $p_\text{top\_center}$ in Fig.~\ref{fig:keypoint_results}. The keypoints can also lie in regions of the object that may be occluded at test time, as exemplified by the $p_\text{bottom\_center}$ keypoint in Fig.~\ref{fig:keypoint_results} which is subject to self-occlusion when the mug is viewed from the side.

Once we have chosen keypoints together with geometric costs and constraints as the problem specification there exist natural formulations for each remaining piece of the manipulation pipeline. This enables us to factor the manipulation policy into 4 subproblems: 1) object instance segmentation 2) category level 3D keypoint detection, 3) a kinematic optimization problem to determine the manipulation action $T_\text{action}$ and 4) grasping the object and executing the desired manipulation action $T_\text{action}$. An illustration of our complete manipulation pipeline is shown in Fig.~\ref{fig:pipeline_figure}. In the following sections, we describe each component of our manipulation pipeline in detail.
%
% For an arbitrary category level manipulation task we can represent an object using task-relevant semantic 3D keypoints. The task is then specified via geometric costs and constraints on these keypoints, which affords a flexible way of formulating the manipulation problem. The user selects keypoints, e.g. $p_\text{top\_center}$ and $p_\text{bottom\_center}$ in the example of Sec.~\ref{subsec:formulation_example}, together with costs and constraints, e.g. (\ref{equ:mug_target_position}) and (\ref{equ:mug_target_axis}), which fully specify the task. Once we have chosen this as the problem specification, there exist natural formulations for each remaining piece of the manipulation pipeline. This allows us to factor the manipulation policy into 4 subproblems: 1) object instance segmentation 2) category level 3D keypoint detection, 3) a kinematic optimization problem to determine the manipulation action $T_\text{action}$ and 4) grasping the object  %(detailed in Sec.~\ref{subsec:robot_details}) 
% and executing the desired manipulation action $T_\text{action}$. An illustration of our complete manipulation pipeline is shown in Fig.~\ref{fig:pipeline_figure}. In the following sections, we describe each component of our manipulation pipeline in detail.

\noindent \textbf{Instance Segmentation and Keypoint Detection } As discussed in Section \ref{subsec:formulation_example} the kPAM pipeline requires being able to detect category-level 3D keypoints from RGBD images of specific object instances. Here we present a specific approach we used to the keypoint detection problem, but note that any technique that can detect these 3D keypoints could be used instead.

We use the state-of-the-art integral network~\cite{sun2018integral} for 3D keypoint detection.
%Need to include instance segmentation
%The network takes a single RGBD image as input. The raw output of the network is the probability heatmap and depth prediction map for each keypoint.
For each keypoint, the network produces a probability heatmap and a depth prediction map as the raw outputs.
The 2-D image coordinates and depth value are extracted using the integral operation~\cite{sun2018integral}. The 3-D keypoints are recovered using the calibrated camera intrinsic parameters. These keypoints are then transformed into world frame using the camera extrinsics. 
% maybe too detailed as part of pipeline
% together with the robot's forward kinematics, as our camera is mounted on the robot's end-effector.

We collect the training data for keypoint detection using a pipeline similar to LabelFusion~\cite{marion2017labelfusion}. Given a scene containing the object of interest we first perform a 3D reconstruction. Then we manually label the keypoints on the 3D reconstruction. We note that this does not require pre-built object meshes. Keypoint locations in image space can be recovered by projecting the 3D keypoint annotations into the camera image using the known camera calibration. Training dataset statistics are provided in Fig.~\ref{fig:quant_results} (c). In total labeling our 117 training scenes took less than four hours of manual annotation time and resulted in over 100,000 labeled images. Even with this relatively small amount of human labeling time we were able to achieve centimeter accurate keypoint detections, enabling us to accomplish challenging tasks requiring high precision, see Section \ref{sec:results}. More details on the keypoint detection network are contained in the supplementary material.

The keypoint detection network~\cite{sun2018integral} requires object instance segmentation as the input, and we integrate Mask R-CNN~\cite{he2017maskrcnn} into our manipulation pipeline to accomplish this step. The training data mentioned above for the keypoint detector~\cite{sun2018integral} can also be used to train the instance segmentation network~\cite{he2017maskrcnn}. Please refer to the supplemental material for more detail.

\noindent \textbf{kPAM Optimization}
The optimization used to find the desired robot action $T_\text{action}^*$ can in general be written as
\begin{equation} \label{equ:general_formulation}
\begin{split}
\underset{T_\text{action} \in SE(3)}{\text{minimize: }}& f(T_\text{action}; \mathbf{p}) \\
\text{subject to: } & \\
 & g(T_\text{action}; \mathbf{p}) = 0 \\ 
 & h(T_\text{action}; \mathbf{p}) \leq 0
\end{split}
\end{equation}
\noindent where $f$ is a scalar cost function, $g$ and $h$ are the equality and inequality constraints, respectively. The robot action $T_\text{action}$ is the decision variable of the optimization problem, and the detected keypoint locations enter the optimization parametrically.

In addition to the constraints used in Sec.~\ref{subsec:formulation_example}, a wide variety of costs and constraints can be used in the optimization~(\ref{equ:general_formulation}). This allows the user to flexibly specify a large variety of manipulation tasks. 
% In practice we found that this specification was rich enough to cover all of our desired use cases. Although an exhaustive list is infeasible.
Below we present several costs/constraints used in our experiments:
\begin{enumerate}
\item L2 distance cost between the transformed keypoint with its nominal target location:
\begin{equation} \label{equ:l2_cost}
 ||T_\text{action}p_i - p_{\text{target}\_i}||^2
\end{equation}
\noindent This is a relaxation of the target position constraint presented in Sec.~\ref{subsec:formulation_example}.
\item Half space constraint on the keypoint:
\begin{equation} \label{equ:halfspace_constraint}
 \langle n_\text{plane}, T_\text{action}p_i \rangle \leq b_\text{plane}
\end{equation}
\noindent where $n_\text{plane} \in R^{3}$ and $b_\text{plane} \in R$ defines the separating plane of the half space. Using the mug in Sec.~\ref{subsec:formulation_example} as an example, this constraint can be used to ensure all the keypoints are above the table to avoid penetration.

\item The point-to-plane distance cost of the keypoint
\begin{equation} \label{equ:pt2pl_cost}
||\langle n_\text{plane}, T_\text{action}p_i\rangle - b_\text{plane}||^2
\end{equation}
\noindent where $n_\text{plane} \in R^{3}$ and $b_\text{plane} \in R$ defines the plane that the keypoint $p_i$ should be in contact with. By using this cost with keypoints that should be placed on the contact surface, for instance the $p_\text{bottom\_center}$ of the mug in Sec.~\ref{subsec:formulation_example}, the optimization (\ref{equ:general_formulation}) can prevent the object from floating in the air.

\item The robot action $T_\text{action}$ should be within the robot's workspace and avoid collisions.
\end{enumerate}

\noindent \textbf{Robot Grasping} Robotic grasping algorithms, such as \cite{gualtieri2016gpd, mahler2019learning}, can be used to apply the abstracted robot action $T_\text{action} \in SE(3)$ produced by the kPAM optimization (\ref{equ:general_formulation}) to the manipulated object. If the object is rigid and the grasp is tight (no relative motion between the gripper and object), applying a rigid transformation to the robot gripper will apply the same transformation to the manipulated object.  These grasping algorithms~\cite{gualtieri2016gpd, mahler2016dex} are object-agnostic and can robustly generalize to novel instances within a given category.

For the purposes of this work we developed a grasp planner which uses the detected keypoints, together with local dense geometric information from a pointcloud, to find high quality grasps. This local geometric information is incorporated with an algorithm similar to the baseline method of \cite{zeng2018affordance}. In general the keypoints used to specify the manipulation task aren't sufficient to determine a good grasp on the object. Thus incorporating local dense geometric information from a depth image or pointcloud can be advantageous. This geometric information is readily available from the RGBD image used for keypoint detection, and doesn't require object meshes. Our grasp planner leverages the detected keypoints to reduce the search space of grasps, allowing us to focus our search on, for example, the heel of a shoe or the rim of a mug. Once we know which aspect of the local geometry to focus on, a high quality grasp can be found by any variety of geometric or learning-based grasping algorithms~\cite{gualtieri2016gpd, mahler2019learning}.

We stress that keypoints are a sparse representation of the object sufficient for describing the manipulation task. However grasping, which depends on the detailed local geoemetry, can benefit from denser RGBD and pointcloud data. This doesn't detract from keypoints as an object representation for manipulation, but rather shows the benefits of different representations for different components of the manipulation pipeline.

\begin{figure}[t]
\centering
\includegraphics[width=0.4\paperwidth]{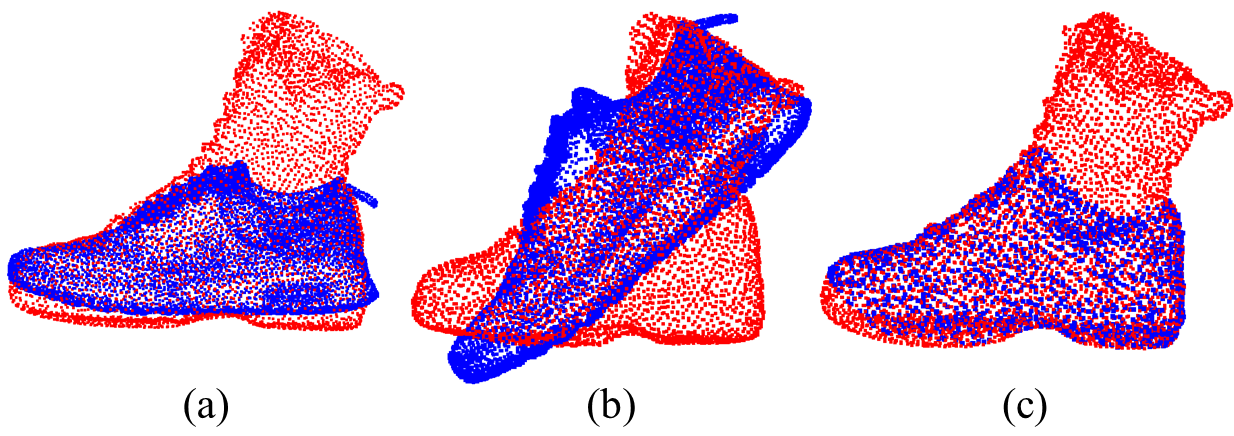}
\caption{\label{fig:ambiguous_align} A pose representation cannot capture large intra-category variations. Here we show different alignment results from a shoe template (blue) to a boot observation (red). (a) and (b) are produced by \cite{gao2019filterreg} with variation on the random seed, and the estimated transformation consists of a rigid pose and a global scale. In (c), the estimated transformation is a fully non-rigid deformation field in~\cite{myronenko2010cpd}. In these examples, the shoe template and transformations can not capture the geometry of the boot observation. Additionally, there may exist multiple suboptimal alignments which make the pose estimator ambiguous. The subsequent robotic pick and place action from these estimations are different, despite these alignments being reasonable geometrically. }
\end{figure}

\begin{figure*}[h]
\centering
\includegraphics[width=1.0\textwidth]{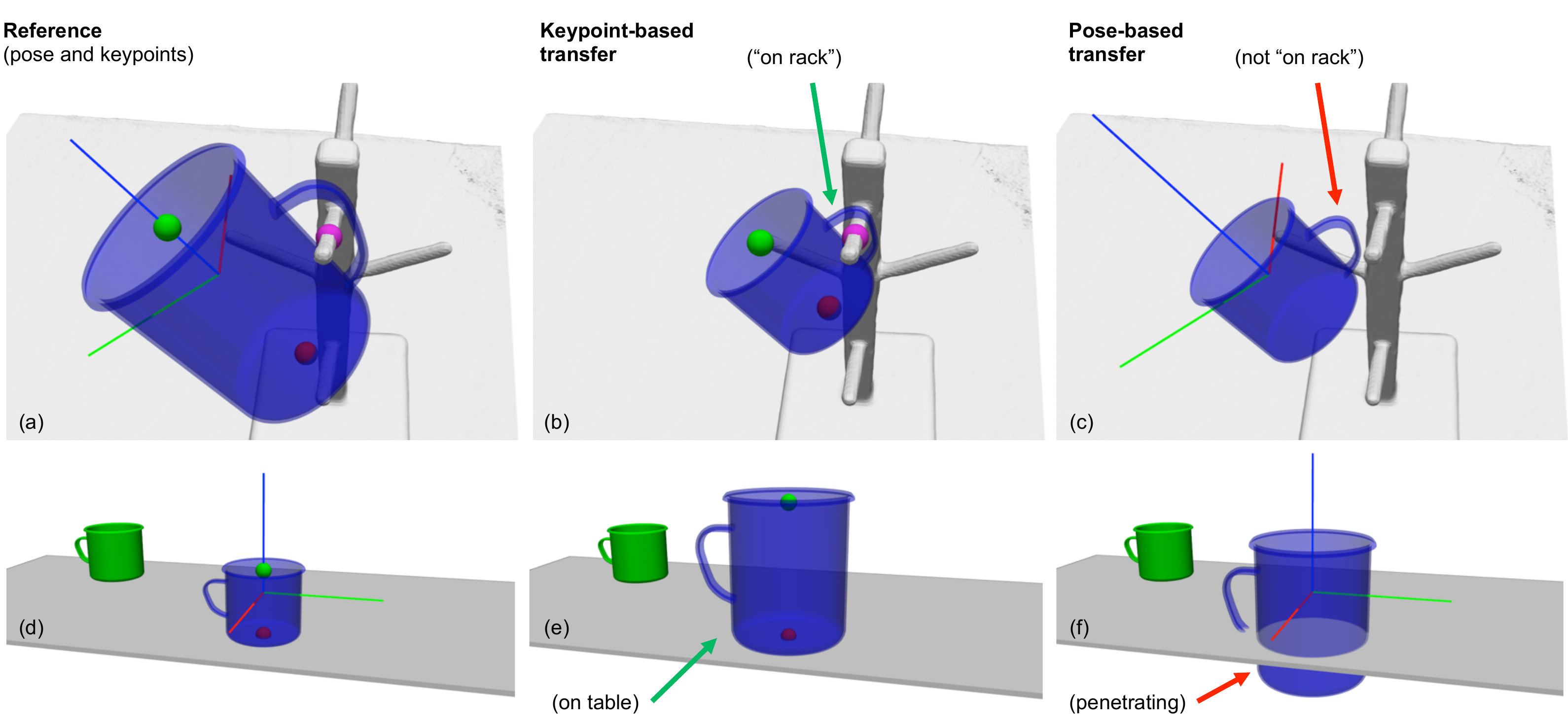}
\caption{\label{fig:director} A comparison of the keypoint based manipulation with pose based manipulation for two different tasks involving mugs. The first row considers the mug on rack task, where a mug must be hung on a rack by its handle. (a) Shows a reference mug in the goal state, (b) and (c) show a scaled down mug instance that could be encountered at test time. (b) uses keypoint based optimization with a constraint on the handle keypoint to find the target state for the mug. The optimized goal state successfully achieves the task of hanging the mug on the rack. In contrast (c) shows the scaled mug instance at the pose defined by (a), which leads to the handle of the mug completely missing the rack, a failure of the task. The second row shows the task of putting a mug on a table. Again (a) shows a reference mug in a goal state, (b) - (c) show a scaled up mug that could be encountered at test time. (b) uses keypoint based optimization with costs/constraints on the bottom and top keypoints to place the mug in a valid goal state. (c) directly uses the pose from (a) on the new mug instance which leads to an invalid goal state where the mug is penetrating the table.}
\end{figure*}

\section{Comparison and Discussions}
\label{sec:comparison}

In this section we compare our approach, as outlined in Sec.~\ref{sec:formulation}, to existing robotic pick and place methods that use pose as the object representation.

\subsection{Keypoint Representation vs Pose Representation}

At the foundation of existing pose-estimation methods is the assumption that the geometry of the object can be represented as a parameterized transformation defined on a fixed template. Commonly used parameterized pose families include rigid, affine, articulated or general deformable. For a given observation (typically an RGBD image or pointcloud), these pose estimators produce a parameterized transformation that aligns the geometric template to the observation.

However, the pose representation is not able to capture large intra-category shape variation. An illustration is presented in Fig.~\ref{fig:ambiguous_align}, where we try to align a shoe template (blue) to a boot observation (red). The alignments in Fig.~\ref{fig:ambiguous_align} (a) and (b) are produced by ~\cite{gao2019filterreg} where the estimated transformation consists of a rigid pose and a global scale. Fig.~\ref{fig:ambiguous_align} (c) is produced by~\cite{myronenko2010cpd} and the estimated transformation is a fully non-rigid deformation field. In these examples, the shoe template and transformations cannot capture the geometry of the boot observation. Additionally, there may exist multiple suboptimal alignments which make the pose estimator ambiguous, as shown in Fig.~\ref{fig:ambiguous_align}. Feeding these ambiguous estimations into a pose-based manipulation pipeline will produce different pick and place actions and final configurations of the manipulated object.

In contrast, we use semantic 3D keypoints as a sparse but task-specific object representation for the manipulation task. Many existing works demonstrate accurate 3D keypoint detection that generalizes to novel instances within the category.
We leverage these contributions to build a robust and flexible manipulation pipeline.

\begin{figure*}[t]
\centering
\includegraphics[width=0.9\textwidth]{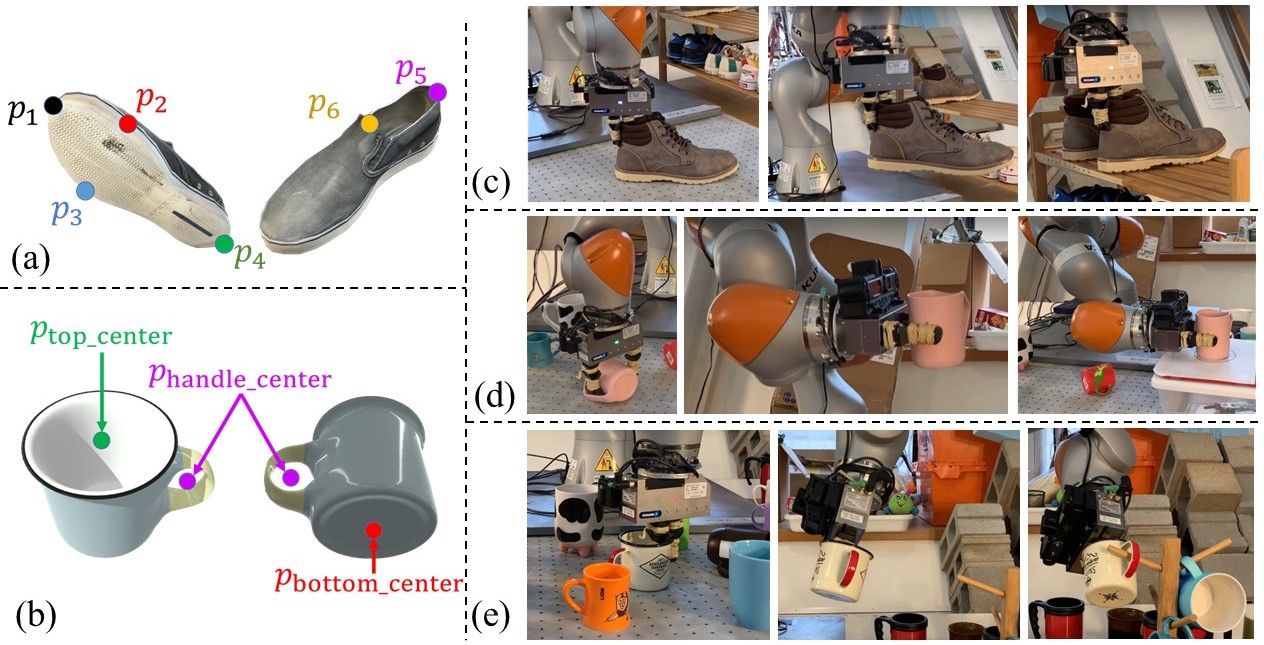}
\caption{\label{fig:keypoint_results} An overview of our experiments. (a) and (b) are the semantic keypoints we used for the manipulation of shoes and mugs. We use three manipulation tasks to evaluate our pipeline: (c) put shoes on a shelf; (d) put mugs on a mug shelf; (e) hang mugs on a rack by the mug handles. The video of these experiments are available on our \href{https://sites.google.com/view/kpam}{\textcolor{blue}{\underline{project page}}}. }
\end{figure*}

Conceptually, a pose representation can also be transformed into keypoint representation given keypoint annotations on the template. However, in practice the transformed keypoints can be inaccurate as the template and the pose cannot fully capture the geometry of new instances. Using the shoe keypoint annotation in Fig.~\ref{fig:keypoint_results} as an example, transforming the keypoints $p_5$ and $p_6$ to a boot using the shoe to boot alignment in Fig~\ref{fig:ambiguous_align} would result in erroneous keypoint detections. A general non-rigid kinematic model (and the associated estimator) that can handle large variations of shape and topology, such as in the example of Fig.~\ref{fig:ambiguous_align}, remains an open problem. Our method avoids this problem by sidestepping the geometric alignment phase and directly detecting the 3D keypoint locations.
%
% Our method can be regarded as generalizations of several existing concepts which attempts to address the limitations of these methods.
%

\subsection{Keypoint Target vs Pose Target}

For existing pose-based pick and place pipelines, the manipulation task is defined as a target pose of the objects. 
For a given scene where the pose of each object has been estimated, these pipelines grasp the object in question and use the robot to move the objects from their current pose to the target pose.
%For a given observation (typically an RGBD image or pointcloud), these methods first estimate the pose of objects in the scene, grasp the object in question and use the robot to move the objects from their current pose to the target pose.

The proposed method can be regarded as a generalization of the pose-based pick and place algorithms. %which can manipulate objects at the category level ('which' is unclear). 
If we detect 3 or more keypoints and assign their target positions as the manipulation goal, then this is equivalent to pose-based manipulation.
In addition, our method can specify more flexible manipulation problems with explicit geometric constraints, such as the bottom of the cup must be on the table and its orientation must be aligned with the upright direction, see Sec. ~\ref{subsec:formulation_example}. The proposed method also naturally generalizes to other objects within the given category, as the keypoint representation ignores many task-irrelevant geometric details. 
%
%On the contrary, pose estimation can be ambiguous under large intra-category variations. Fig.~\ref{fig:ambiguous_align} provides an illustrative example using different alignment results between a shoe and a boot. Fig.~\ref{fig:ambiguous_align} (a) and (b) are produced by ~\cite{gao2019filterreg} where the estimated transformation consists of a rigid pose and a global scale. Fig.~\ref{fig:ambiguous_align} (c) is produced by~\cite{myronenko2010cpd} and the estimated transformation is the fully nonrigid deformation field. Suppose the target pose is defined for the shoe, existing pose-based manipulation pipelines will produce different pick and place actions and final configurations for the boot given the alignment results in Fig.~\ref{fig:ambiguous_align}, despite all alignments being quite reasonable geometrically.
%
%
%Additionally, defining a target pose at the category level can lead to target manipulation actions that are physically infeasible. 

On the contrary a pose target is object-specific and defining a target pose at the category level can lead to manipulation actions that are physically infeasible.
Consider the mug on table task from Section \ref{subsec:formulation_example}. Fig.~\ref{fig:director} (d) shows the target pose for the reference mug model. Directly applying this pose to the scaled mug instance in Fig. ~\ref{fig:director} (f) leads to physically infeasible state where the mug is penetrating the table. In contrast, using the optimization formulation of Section \ref{sec:formulation} results in the mug resting stably on the table, shown in Fig.~\ref{fig:director} (e).

In addition to leading to states which are physically infeasible, pose-based targets at a category level can also lead to poses which are physically feasible but fail to accomplish the manipulation task. Figures \ref{fig:director} (a) - (c) show the mug on rack task. In this task the goal is to hang a mug on a rack by its handle. Fig.~\ref{fig:director} (a) shows the reference model in the goal state. Fig.~\ref{fig:director} (c) shows the result of applying the pose based target to the scaled down mug instance. As can be seen even though the pose unambiguously matches the target pose exactly, this state doesn't accomplish the manipulation task since the mug handle completely misses the rack. Fig.~\ref{fig:director} (b) shows the result of our kPAM approach. Simply by adding a constraint that handle center keypoint should be on the rack, a valid goal state is returned by the kPAM optimization.

\section{Results}
\label{sec:results}

In this section, we demonstrate a variety of pose-aware pick and place tasks using our keypoint-based manipulation pipeline. The particular novelty of these demonstrations is that our method is able to handle large intra-category variations without any instance-wise tuning or specification. We utilize a 7-DOF robot arm (Kuka IIWA LBR) mounted with a Schunk WSG 50 parallel jaw gripper. An RGBD sensor (Primesense Carmine 1.09) is also mounted on the end effector. The video demo on our \href{https://sites.google.com/view/kpam}{\textcolor{blue}{\underline{project page}}} best demonstrates our solution to these tasks. More details about the experimental setup are included in the supplemental material.

% \begin{figure}[t]
% \centering
% \includegraphics[width=0.45\textwidth]{figures/quant_results_vertical_crop.pdf}
% \caption{\label{fig:quant_results} Quantitative results from the 3 hardware experiments. (a) and (b) show some of the test objects for the experiments. (c) statistics of the training data (d) We report the average heel and toe errors (along the horizontal direction) from their desired locations as well as the standard deviation. (e) The reported errors for the mug on shelf task are the distance from the bottom center keypoint to the target location of that keypoint in the optimization program. (f) reports success rates for the mug on rack task for different sized mugs. Mugs with handles having either height or width less than 2cm are classified as ``small'' (more details in supplementary material). A trial was deemed successful if the mug ended up hanging on the rack by the mug handle. Videos of the experiments are available on our \href{https://sites.google.com/view/kpam}{\textcolor{blue}{\underline{project page}}}.} 
% \end{figure}

\begin{figure}[t]
\centering
\includegraphics[width=0.9\textwidth]{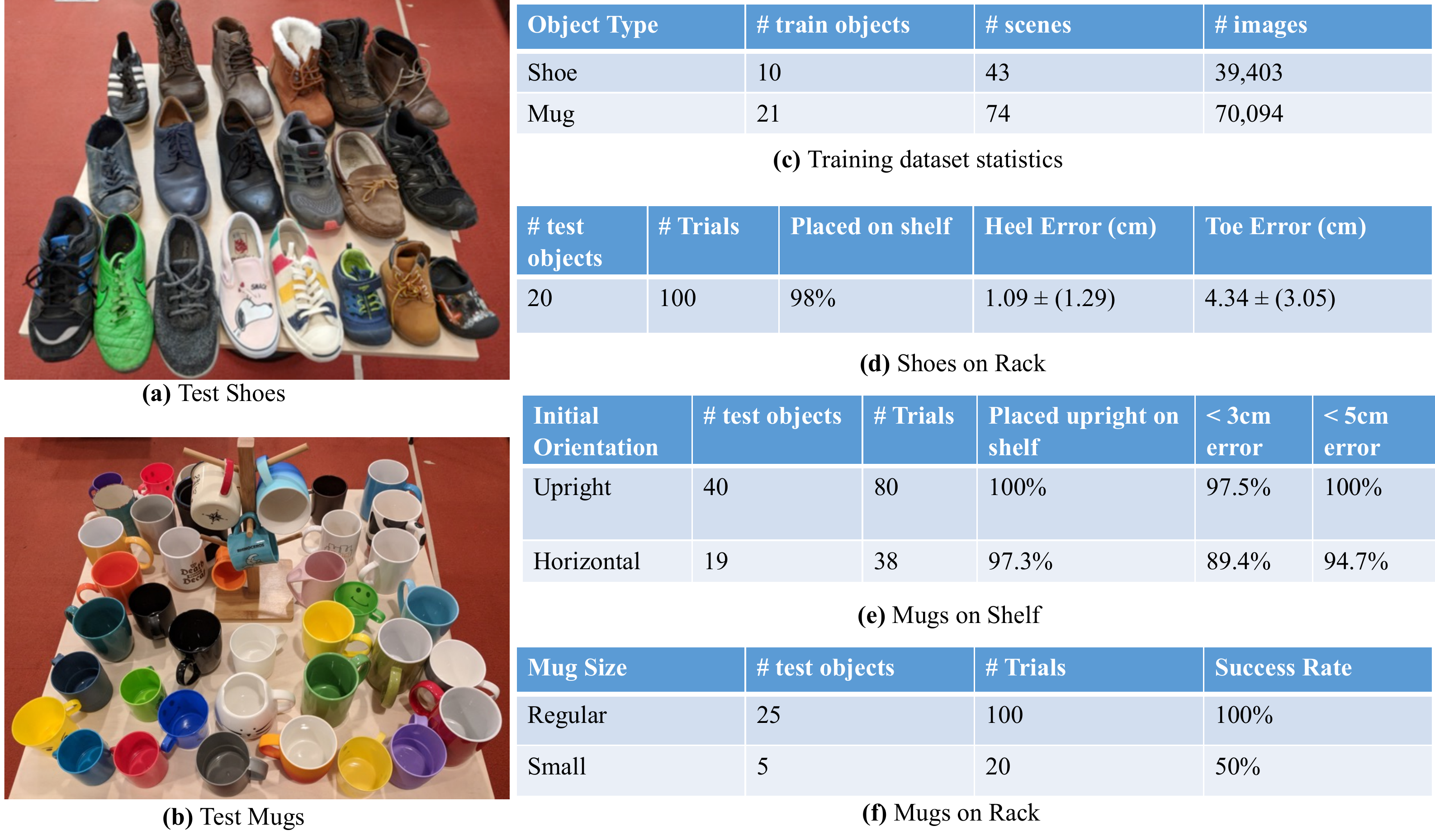}
\caption{\label{fig:quant_results} Quantitative results from the 3 hardware experiments. (a) and (b) show some of the test objects for the experiments. (c) statistics of the training data (d) We report the average heel and toe errors (along the horizontal direction) from their desired locations as well as the standard deviation. (e) The reported errors for the mug on shelf task are the distance from the bottom center keypoint to the target location of that keypoint in the optimization program. (f) reports success rates for the mug on rack task for different sized mugs. Mugs with handles having either height or width less than 2cm are classified as ``small'' (more details in supplementary material). A trial was deemed successful if the mug ended up hanging on the rack by the mug handle. Videos of the experiments are available on our \href{https://sites.google.com/view/kpam}{\textcolor{blue}{\underline{project page}}}.} 
\end{figure}

\subsection{Put shoes on a shoe rack}

\subsubsection{Task Description}

\noindent \textbf{Task Description } Our first manipulation task is to put shoes on a shoe rack, as shown in Fig.~\ref{fig:keypoint_results} (c). We use shoes with different appearance and geometry to evaluate the generality and robustness of our manipulation policy. The six keypoints used in this manipulation task are illustrated in Fig~\ref{fig:keypoint_results} (a), and the costs and constraints in the optimization (\ref{equ:general_formulation}) are

\begin{enumerate}
\item The L2 distance cost (\ref{equ:l2_cost}) between keypoints $p_1$, $p_2$, $p_3$ and $p_4$ to their nominal target locations.
\item The sole of the shoe should be in contact with the rack surface. In particular, the point-to-plane cost (\ref{equ:pt2pl_cost}) is used to penalize the deviation of keypoints $p_2$, $p_3$ and $p_4$ from the supporting surface.
\item All the keypoints should be above the supporting surface to avoid penetration. A half-space constraint (\ref{equ:halfspace_constraint}) is used to enforce this condition.
\end{enumerate}

For our experiments we place the shoe rack in a known position, but this constraint could be easily relaxed by adding a pose-estimation module for the shoe-rack.

\noindent \textbf{Experimental Results } The shoe keypoint detection network was trained on a labeled dataset of 10 shoes, detailed in Figure \ref{fig:quant_results} (c). Experiments were conducted with a held out test set of 20 shoes with large variations in shape, size and visual appearance (more details in the video and supplemental material). For each shoe we ran 5 trials of the manipulation task. Each trial consisted of a single shoe being placed on the table in front of the robot. Using the kPAM pipeline the robot would pick up the shoe and place it on a shoe rack. The shoe rack was marked so that the horizontal deviation of the shoe's toe and heel bottom keypoints ($p_1$ and $p_4$ respectively in Fig.~\ref{fig:keypoint_results}) from their nominal target locations %as specified in the optimization program, 
could be determined. 
Quantitative results are given in Fig. \ref{fig:quant_results} (d). Out of 100 trials only twice did the pipeline fail to place the shoe on the rack. Both failures were due to inaccurate keypoint detections. One led to a failed grasp and another to an incorrect $T_\text{action}$. For trials which ended up with the shoe on the rack average errors for the heel and toe keypoint locations are given in Fig.~\ref{fig:quant_results} (d). During the course of our experiments we noticed that the majority of these errors come from the fact that when the robot grasps the shoe by the heel the closing of the gripper often results in the object shifting from the position it was in when the RGBD image used for keypoint detection was captured. This accounts for the majority of the errors observed in the final heel and toe keypoint locations. The keypoint detections and resulting $T_\text{action}$ would have almost always results in heel and toe errors of less than 1 cm if we were able to exactly apply $T_\text{action}$ to the object. Since our experimental setup relies on a wrist mounted camera we are not able to re-perceive the object after grasping it. We believe that these errors could be further reduced by adding an external camera that would allow us to re-run our keypoint detection after grasping the object to account for any object movement during the grasp. Overall kPAM approach was very successful at the shoes on rack task with a greater than 97\% success rate.

% \vspace{-1em}
\subsection{Put mugs upright on a shelf}
% \vspace{-1em}
\label{subsec:mugs_on_shelf}

\noindent \textbf{Task Description } We also perform a real-world demonstration of the ``put mugs upright on a shelf" task described in Sec.~\ref{subsec:formulation_example}, as shown in Fig.~\ref{fig:keypoint_results} (d). The keypoints used in this task are illustrated in Fig.~\ref{fig:keypoint_results} (b). The costs and constraints for this task include the target position constraint~(\ref{equ:mug_target_position}) and the axis alignment constraint~(\ref{equ:mug_target_axis}). This task is very similar to the mugs task in \cite{gualtieri2018pick}. For this set of experiments we didn't place any costs or constraints on the yaw degree of freedom of the mug, but if a specific yaw orientation was desired this could be incorporated by adding an L2 cost~(\ref{equ:l2_cost}) between the $p_\text{handle\_center}$ keypoint with its target location. 

\noindent \textbf{Experimental Results } The mug keypoint detection network was trained on a dataset of $21$ standard sized mugs, detailed in Fig.~\ref{fig:quant_results} (c). Experiments for the mug on shelf task were conducted using a held out test set of 40 mugs with large variations in shape, size and visual appearance (more details in the video and supplemental material). All mugs could be grasped when in the upright orientation, but due to the limited stroke of our gripper (7.5cm when fully open) only 19 of these mugs could be grasped when lying horizontally. For mugs in that could be grasped horizontally we ran two trials with the mug starting from a horizontal orientation, and two trials with the mug in a vertical orientation. For the remaining mugs we ran two trials for each mug with the mug starting in an upright orientation. Quantitative performance was evaluated by recording whether the mug ended up upright on the shelf, and the distance of the mug's bottom center keypoint to the target location. Results are shown in Fig.~\ref{fig:quant_results} (e). Overall our system was very reliable, managing to place the mug on the shelf within 5cm of the target location in all but 2 trials. In one of these failures the mug was placed upside down. In this case the mug was laying horizontally on the table and the RGB image used in keypoint detection (see  Fig.~\ref{fig:mug_upside_down} in the Appendix) was taken from a side-on profile where the handle is occluded and it is very difficult to distinguish the top from the bottom of the mug. This led our keypoint detector to mix up the top and bottom of the mug, causing it to be placed upside down. The keypoint detection error is understandable in this case since it is very difficult to distinguish the top from the bottom of this mug in the single RGBD image. In addition this particular instance was a small kids sized mug, whereas all the training data for mugs contained only regular sized mugs. See Section \ref{sec:limitations} for more discussion on this failure.

% Fig.~\ref{fig:mug_upside_down} shows the RGB image used in keypoint detection along with the final object placement. In this case the keypoint detection mixed up the top and bottom of the mug, causing it to be placed upside down. The keypoint detection error is understandable in this case since it is very difficult to distinguish the top from the bottom of this mug in the single RGBD image. In addition this particular instance was a small kids sized mug, whereas all the training data for mugs contained only regular sized mugs.
% % See Section \ref{sec:limitations} for more discussion on this failure.
%
Overall the accuracy in the mug on shelf task was very high, with $97\%$ of upright trials, and $88\%$ of horizontal trials resulting in bottom keypoint final location errors of less than 3cm. Qualitatively the majority of this error arose from the object moving slightly during the grasping process with the rest attributed to the keypoint detection.
%

% \begin{figure}[t]
% \centering
% \includegraphics[width=0.45\textwidth]{figures/mug_upside_down_crop.pdf}
% \caption{\label{fig:mug_upside_down} (a) The RGB image for the single failure trial of the mug on shelf task that led to the mug being put in an incorrect orientation. In this case the keypoint detection confused the top and bottom of the mug and it was placed upside down. (b) The resulting upside down placement of the mug.}
% \end{figure}
%
%
\subsection{Hang the mugs on the rack by their handles}
\label{subsec:mug_rack_experiments}
\noindent \textbf{Task Description } To demonstrate the accuracy and robustness of our method we tasked the robot with autonomously hanging mugs on a rack by their handle. An illustration of this task is provided in Fig.~\ref{fig:keypoint_results} (e). The relatively small mug handles (2-3 centimeters) challenge the accuracy of our manipulation pipeline. The costs and constraints in this task are
\begin{enumerate}
\item The target location constraint (\ref{equ:mug_target_position}) between $p_\text{handle\_center}$ to its target location on the rack axis.
\item The keypoint L2 distance cost (\ref{equ:l2_cost}) from $p_\text{top\_center}$ and $p_\text{bottom\_center}$ to their nominal target locations.
\end{enumerate}
In order to avoid collisions between the mug and an intermediate goal for the mug was specified. Using the notation of (\ref{equ:general_formulation}) this intermediate goal $T_\text{approach}$ was gotten by shifting $T_\text{action}$ away from the rack by 10cm along the direction of the target peg. We then executed the final placement by moving the end effector in a straight line connecting $T_\text{approach}$ to $T_\text{action}$. During these experiments the mug-rack was placed in a fixed known position, but this constraints be easily relaxed by adding a pose-estimation module for the mug-rack.

\noindent \textbf{Experimental Results } For the mug on rack experiments we used the same keypoint detection network as for the mug on shelf experiments. Experiments were conducted using a held out test set of 30 mugs with large variation in shape, texture and topology. Of these 5 were very small mugs whose handles had a minimum dimension (either height or width) of less than 2cm (see the supplementary material for more details). We note that the training data did not contain any such ``small'' mugs. Each trial consisted of placing a single mug on the table in front of the robot. Then the kPAM pipeline was run and a trial was recorded as successful if the mug ended up hanging on the rack by its handle. Five trials were run for each mug and quantitative results are reported in Fig.~\ref{fig:quant_results} (e). For regular sized mugs we were able to hang them on the rack with a 100\% success rate. The small mugs were much more challenging but we still achieved a 50\% success rate. The small mugs have very tiny handles, which stresses the accuracy of the entire system. In particular the total error of the keypoint detection, grasping and execution needed to successfully complete the task for the small mugs was on the order of 1-1.5 cm. Two main factors contributed to failures in the mug on rack task. The first, similar to the case of shoe on rack task, is that during grasping the closing of the gripper often moves the object from the location at which it was perceived. Even a small disturbance (i.e. $<$ 1cm) can lead to a failure in the mug on rack task since the required tolerances are very small. The second contributing factor to failures is inaccurate keypoint detections. Again an inaccurate detection of even 0.5-1cm can be sufficient for the mug handle to miss the rack entirely. As discussed previously, the movement of the object during grasping could be alleviated by the addition of an external camera that would allow us to re-perceive the object after grasping.

% \vspace{-1em}
\section{Limitations and Future Work}
\label{sec:limitations}

Our current data collection pipeline in Sec.~\ref{subsec:general_formulation} requires human annotation, although the use of 3D reconstruction somewhat alleviates this manual labor. An interesting direction for future work is to train our keypoint detector using synthetic data, as demonstrated in~\cite{tremblay2018deep, wang2019normalized}.

Representing the robot action with a rigid transformation $T_\text{action}$ is valid for robotic pick-and-place. However, this abstraction does not work for deformable objects or more dexterous manipulation actions on rigid objects, such as the in-hand manipulation in~\cite{andrychowicz2018learning}. Combining these learning-based or model-based approaches with the keypoint representation to build a manipulation policy that generalizes to categories of objects would is a promising direction for future work. In addition to the usage in our pipeline, we believe that the keypoint representation can potentially contribute to various learning-based manipulation approaches as 1) a reward function to flexibly specify the manipulation target or 2) an alternative input to the policy/value neural network, which is more robust to shape variation and large deformation than the widely-used pose representation.

% \vspace{-1em}
\section{Conclusion}
% \vspace{-1em}
\label{sec:conclusion}
In this paper we contribute a novel formulation of category-level manipulation which uses semantic 3D keypoints as the object representation. Using keypoints to represent the object enables us to simply and interpretably specify the manipulation target as geometric costs and constraints on the keypoints, which flexibly generalizes existing pose-based manipulation methods. This formulation naturally allows us to factor the manipulation policy into the 3D keypoint detection, optimization-based robot action planning and grasping based action execution. By factoring the problem we are able to leverage advances in these sub-problems and combine them into a general and effective perception-to-action manipulation pipeline. Through extensive hardware experiments, we demonstrate that our pipeline is robust to large intra-category shape variation and can accomplish manipulation tasks requiring centimeter level precision.
%
% \vspace{-1em}
\section*{Acknowledgements}
% \vspace{-0.5em}
The authors thank Ethan Weber (instance segmentation training data generation) and Pat Marion (visualization) for their help. This work was supported by: National Science Foundation, Award No. IIS-1427050; Draper Laboratory Incorporated, Award No. SC001-0000001002;  Lockheed Martin Corporation, Award No. RPP2016-002; Amazon Research Award.
{\small
% %\bibliographystyle{styles/bibtex/spmpsci.bst}
% \bibliographystyle{main}{abbrv}
% \bibliography{main}{paper.bib}{References}
\bibliographystyle{plain}
\bibliography{paper.bib}
}

\newpage

% \section{Supplementary Material}
\appendix

\section{Robot Hardware}
Our experimental setup consists of a robot arm, an end-effector mounted RGBD camera and a parallel jaw gripper. Our robot is a 7-DOF Kuka IIWA LBR. Mounted on the end-effector is a Schunk WSG 50 parallel jaw gripper. Additionally we mount a Primesense Carmine 1.09 RGBD sensor to the gripper body.

\section{Dataset Generation and Annotation}

In order to reduce the human annotation time required for neural network training we use a data collection pipeline similar to that used in \cite{florence2018dense}. The main idea is to collect many RGBD images of a static scene and perform a dense 3D reconstruction. Then, similarly to \cite{marion2017labelfusion}, we can label the 3D reconstruction and propagate these labels back to the individual RGBD frames. This 3D to 2D labelling approach allows us to generate over 100,000 labelled images with only a few hours of human annotation time.

% in the 3D reconstruction and similar approach uses the wrist mounted camera to  In This approach uses the wrist-mounted camera to perform a dense 3D reconstruction of the scene. We then label this 3D reconstruction and propagate the labels to the individual RGBD images. A description of this approach, taken from \cite{florence2018dense}, is reproduced here for completeness.

\subsection{3D Reconstruction and Masking}
\label{subsec:3D_reconstruction}

Here we give a brief overview of the approach used to generate the 3D reconstruction, more details can be found in \cite{florence2018dense}. Our data is made up of 3D reconstructions of a static scene containing a single object of interest. Using our the wrist mounted camera on the robot, we move the robot's end-effector to capture a variety of RGBD images of the static scene. From the robot's forward kinematics, we know the camera pose corresponding to each image which allows us to use TSDF fusion \cite{curless1996volumetric} to obtain a dense 3D reconstruction. After discarding images that were taken from very similar poses, we are left with approximately 400 RGBD images per scene. 

The next step is to detect which parts of the 3D reconstruction correspond to the object of interest. This is done using the change detection method described in \cite{finman2013toward}. In our particular setup all the reconstructions were of a tabletop scene in front of the robot. Since our reconstructions are globally aligned (due to the fact that we use the robot's forward kinematics to compute camera poses), we can simply crop the 3D reconstruction to the area above the table. At this point we have the portion of the 3D reconstruction that corresponds to the object of interest. This, together with the fact that we have camera poses, allows us to easily render binary masks (which segments the object from the background) for each RGBD image.

\subsection{Instance Segmentation}
\label{subsec:instace_segmentation_data}

\begin{figure}[!tbp]
\centering
  \begin{subfigure}[b]{0.4\textwidth}
    \includegraphics[width=\textwidth]{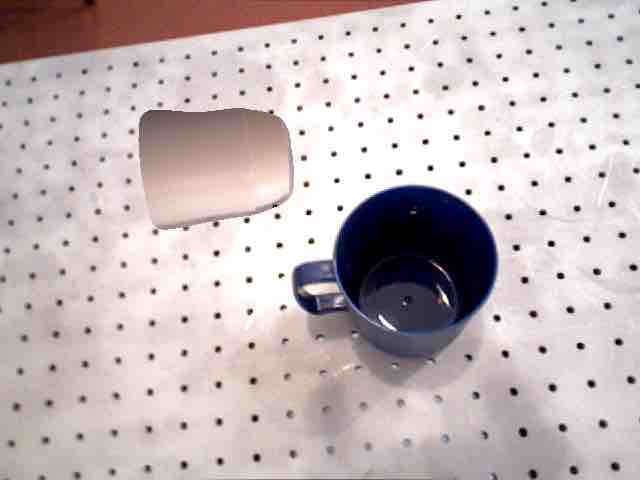}
    \caption{Mugs composite image}
    \label{fig:shoes_composite}
  \end{subfigure}
  \hfill
  \begin{subfigure}[b]{0.4\textwidth}
    \includegraphics[width=\textwidth]{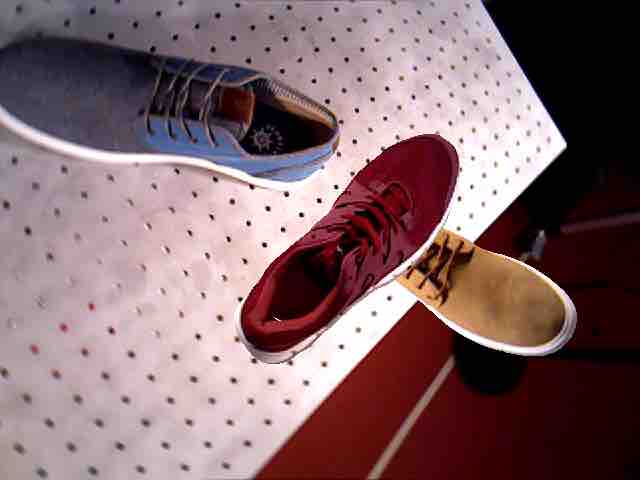}
    \caption{Shoes composite image}
    \label{fig:mugs_composite}
  \end{subfigure}
  \caption{Multi object composite images used in instance segmentation training}
  \label{fig:composite_images}
\end{figure}

The instance segmentation network requires training images with pixelwise semantic labels. Using the background subtraction technique detailed in Section~\ref{subsec:3D_reconstruction}, we have pixelwise labels for all the images in our 3D reconstructions. However, these images contain only a single object, while we need the instance segmentation network to handle multiple instances at the test time. %Similar to \cite{schwarz2018fast} we augment our training data by creating multi-object composite images from our single object annotated images. 
Thus, we augment the training data by creating multi-object composite images from our single object annotated images using a method similar to \cite{schwarz2018fast}. We crop the object from one image (using the binary mask described in Section~\ref{subsec:3D_reconstruction}) and paste this cropped section on top of an existing background.
%This is done by simply pasting the cropped part of the image corresponding to the object in question on top of an existing background. 
This process can be repeated to generate composite images with arbitrary numbers of object. %For our experiments, we generated images with between 1 and 7 objects. 
Examples of such images are shown in Figure \ref{fig:composite_images}.

\subsection{Keypoint Detection}
\label{subsec:keypoint_data}

\begin{figure}[t]
\centering
\includegraphics[width=0.6\textwidth]{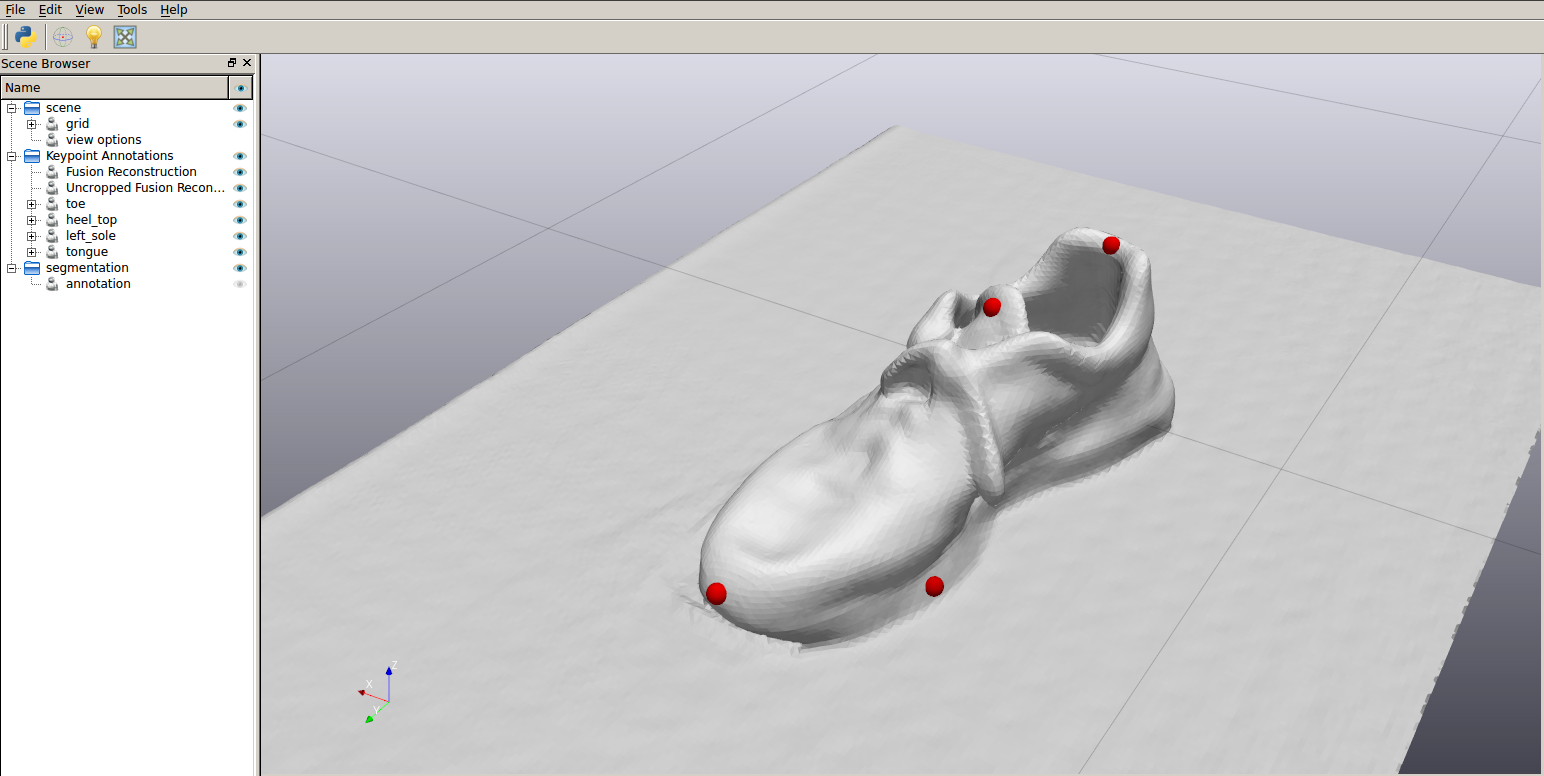}
\caption{\label{fig:director_keypoint_annotation} A screenshot from our custom keypoint annotation tool.}
\end{figure}

The keypoint detection network requires training images annotated with pixel coordinates and depth for each keypoint. As mentioned in Section~\ref{subsec:general_formulation}, we annotate 3D keypoints on the reconstructed mesh, transform the keypoints into the camera frame and project the keypoints into each image. To annotate the 3D keypoints on the reconstructed mesh, we developed a custom labelling tool based on the Director \cite{director} user interface, shown in Figure~\ref{fig:director_keypoint_annotation}. 
%As detailed in Figure \ref{fig:quant_results},
We labelled a total of 117 scenes, 43 of which were shoes and 74 of which were mugs. Annotating these scenes took only a few hours and resulted in over 100,000 labelled images for keypoint network training.

%We developed a custom labelling tool based on the Director \cite{director} visualizer for annotating keypoint locations in the 3D reconstruction. Using this tool annotating a 3D reconstruction takes on the order of 1-2 minutes depending on the number of keypoints to be labelled (3 for mugs, 6 for the shoes). Once the 3D reconstruction is annotated we can backproject these labels into each of the 400 individual RGBD frames corresponding to that scene. The fact that we only have to annotate the 3D reconstruction vastly improves the efficiency of our data generation and labelling approach. As detailed in Figure \ref{fig:quant_results} we labelled a total of 117 scenes, 43 of which were shoes and 74 of which were mugs. Annotating these scenes took only a few hours and resulted in over 100,000 labelled images.
%

\section{Neural Network Architecture and Training}

\subsection{Instance Segmentation}

For the instance segmentation, we used an open source Mask R-CNN implementation \cite{massa2018mrcnn}. We used a R-101-FPN backbone that was pretrained on the COCO dataset \cite{lin2014microsoft}. We then fine-tuned on a dataset of 10,000 images generated using the procedure outlined in Section \ref{subsec:instace_segmentation_data}. The network was trained for 40,000 iterations using the default training schedule of \cite{massa2018mrcnn}.

\begin{figure}[!tbp]
\centering
  \begin{subfigure}[b]{0.6\textwidth}
  \centering
    \includegraphics[width=\textwidth]{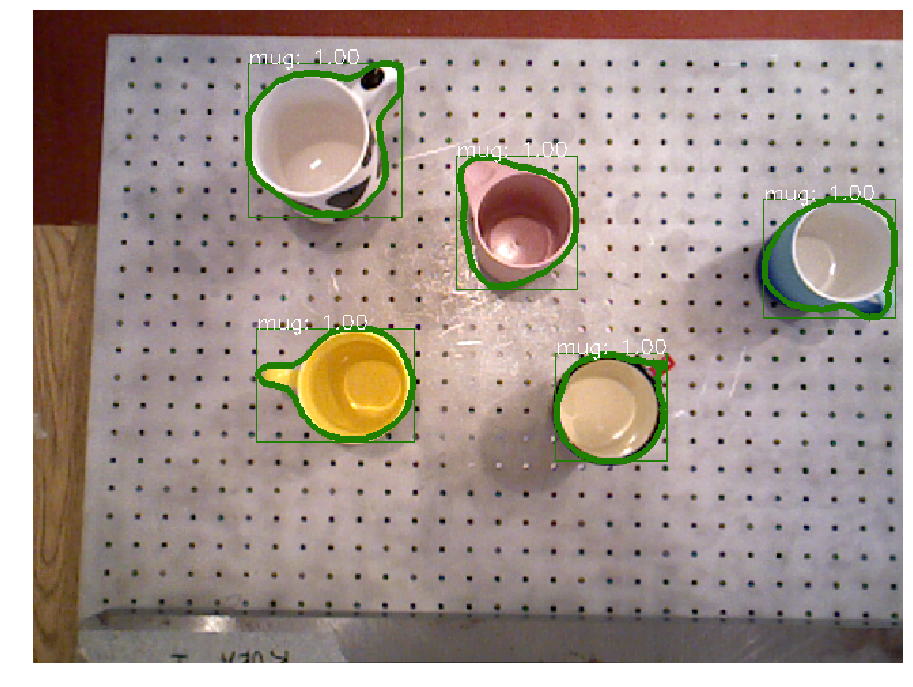}
    \caption{RGB image used for keypoint detections with Mask R-CNN annotations overlaid}
    \label{fig:multiple_mugs_rgb}
  \end{subfigure}
  \hfill
  \begin{subfigure}[b]{0.6\textwidth}
  \centering
    \includegraphics[width=\textwidth]{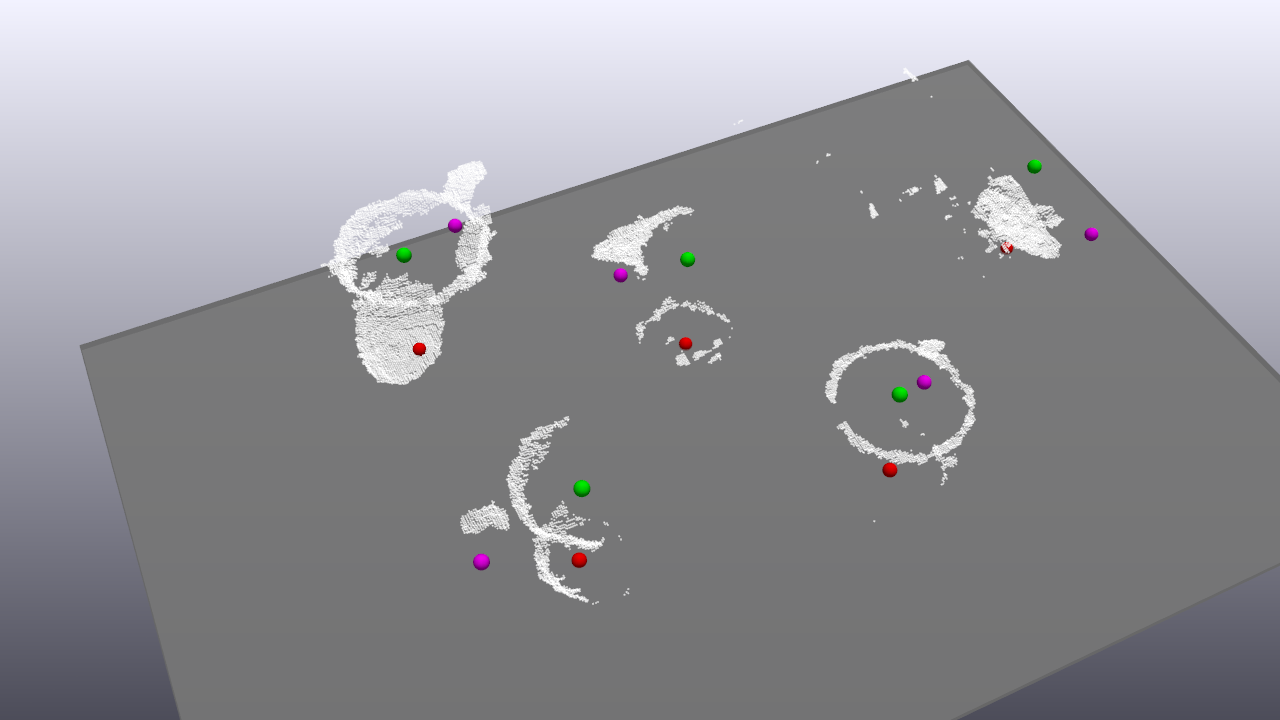}
    \caption{Keypoint detections}
    \label{fig:multiple_mugs_keypoint}
  \end{subfigure}
  \caption{3D visualization of pointcloud and keypoint detections for the image from (a). The keypoints are colored as in Figure \ref{fig:keypoint_results}. The \emph{top center} keypoint is green, the \emph{bottom center} keypoint is red, and the \emph{handle center} keypoint is purple.}
  \label{fig:keypoint_detection_details}
\end{figure}

\subsection{Keypoint Detection}

We modify the integral network~\cite{sun2018integral} for 3D keypoint detection. The network takes images cropped by the bounding box from MaskRCNN as the input. The network produces the probability distribution map $g_i(u, v)$ that represents how likely keypoint $i$ is to occur at pixel $(u,v)$, with $\sum_{u,v}g_i(u,v)=1$. We then compute the expected values of these spatial distributions to recover a pixel coordinate of the keypoint $i$:

\begin{equation}
    [u_i, v_i]^T=\sum_{u, v} [u \cdot g_i(u,v), v \cdot g_i(u,v)]^T
\end{equation}

For the $z$ coordinates (depth) of the keypoint, we also predict a depth value at every pixel denoted as $d_i(u,v)$. The depth of the keypoint $i$ can be computed as

\begin{equation}
    z_i=\sum_{u, v} d_i(u,v) \cdot g_i(u,v)
\end{equation}

Given the training images with annotated pixel coordinate and depth for each keypoint, we use the integral loss and heatmap regression loss (see Section 2 of \cite{sun2018integral} for details) to train the network. We use a network with a 34 layers Resnet as the backbone. The network is trained on a dataset generated using the procedure described in Section~\ref{subsec:keypoint_data}.

\section{Experiments}

\begin{figure}[!tbp]
\centering
  \begin{subfigure}[b]{1.0\textwidth}
  \centering
    \includegraphics[trim=150 150 150 150,clip,width=\textwidth]{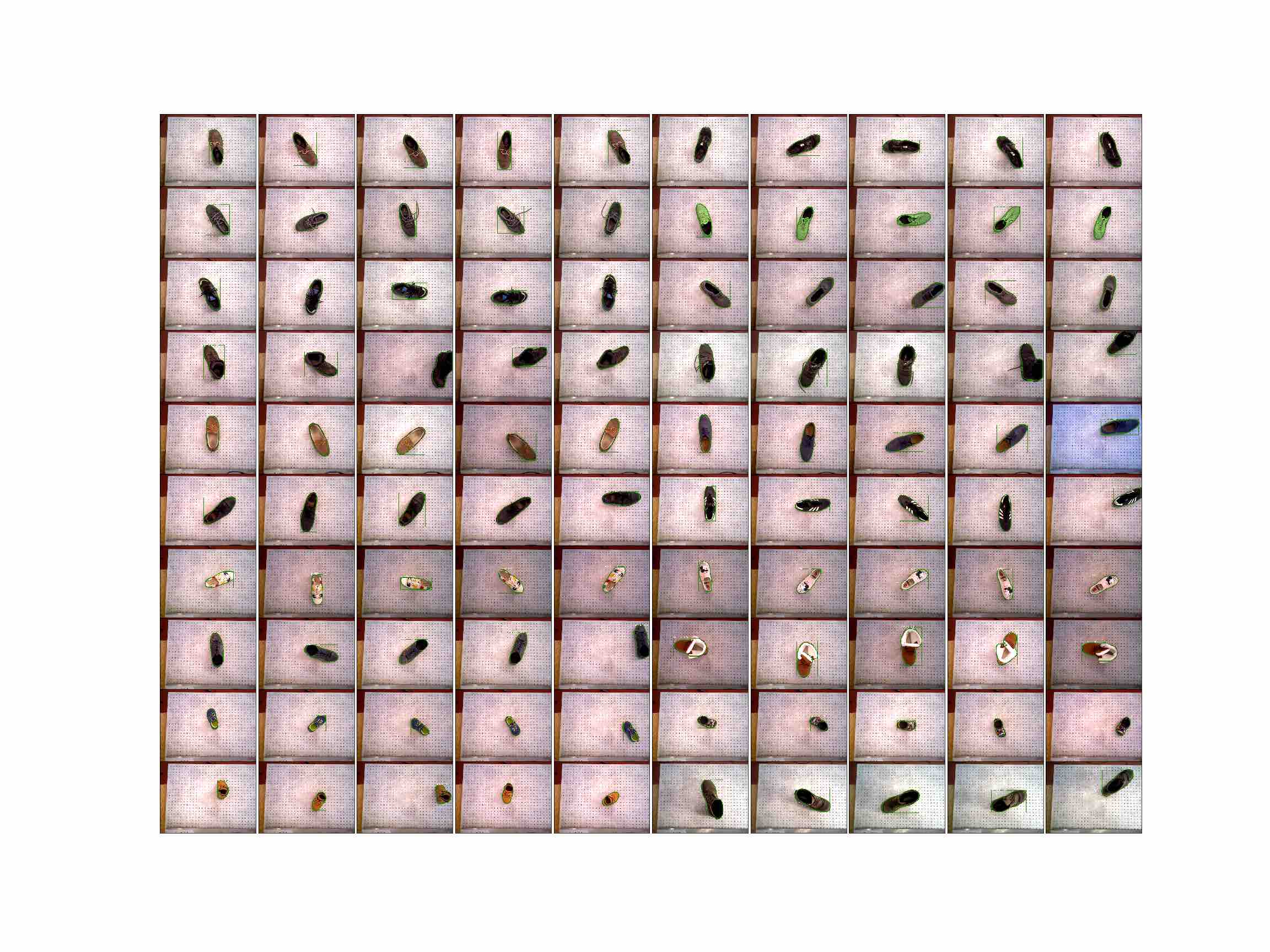}
    \caption{RGB images used for keypoint detection in shoe on rack experiments}
    \label{fig:shoes_tiled_start_pose}
  \end{subfigure}
  \hfill
  \begin{subfigure}[b]{1.0\textwidth}
  \centering
    \includegraphics[trim=100 100 100 100,clip,width=\textwidth]{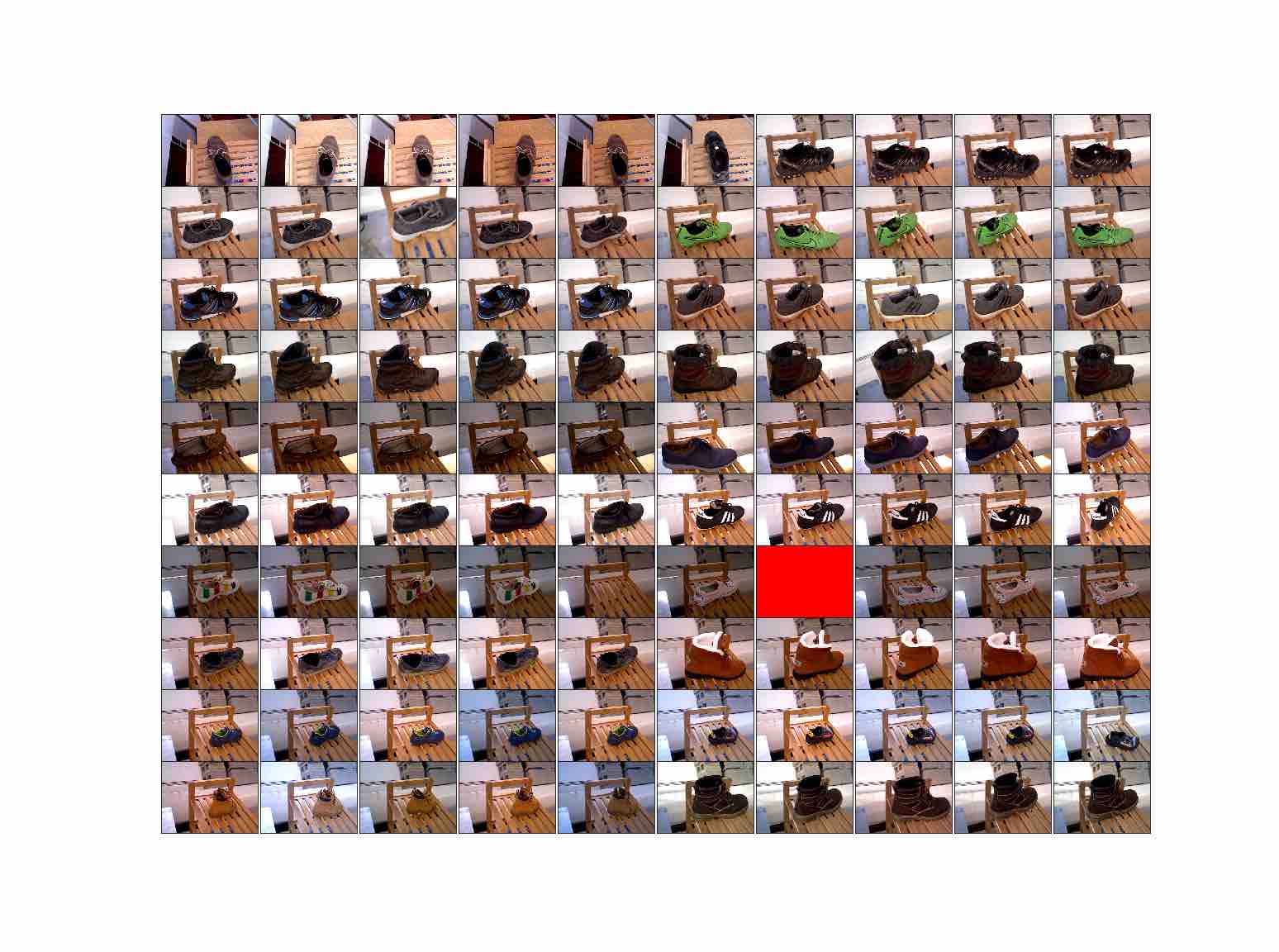}
    \caption{Image of the shoe rack after running the kPAM pipeline. Red images indicate trials where no image of the final placement was captured due to an upstream failure of the pipeline causing the trial to be aborted.}
    \label{fig:shoes_tiled_end_pose}
  \end{subfigure}
  \caption{Before and after images of the shoe on rack experiment for all 100 trials.}
  \label{fig:composite_images_shoe_on_shelf}
\end{figure}

\begin{figure}[!tbp]
\centering
  \begin{subfigure}[b]{1.0\textwidth}
  \centering
    \includegraphics[trim=150 150 150 150,clip,width=\textwidth]{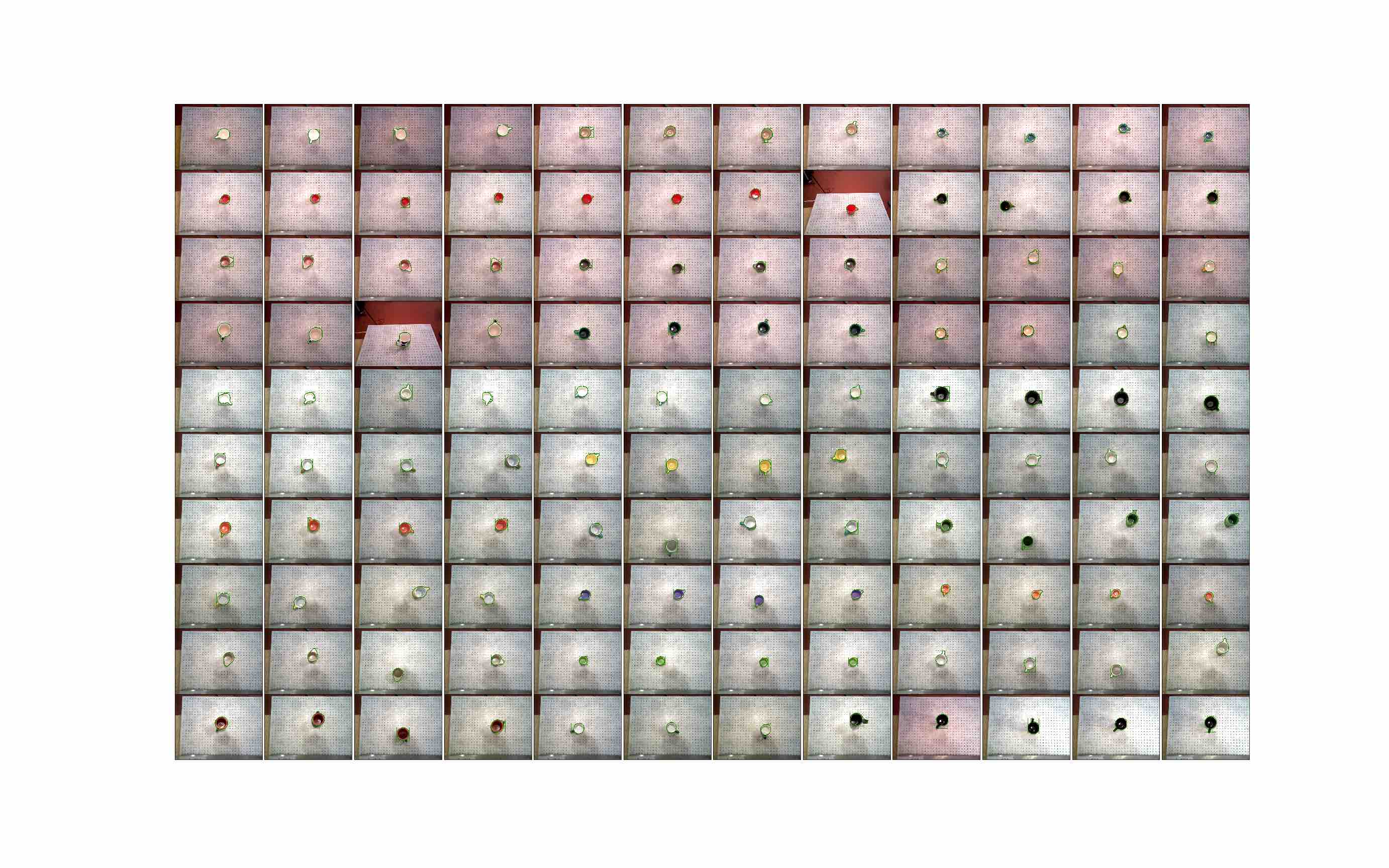}
    \caption{RGB images used for keypoint detection in mug on rack experiments}
    \label{fig:mug_on_rack_tiled_start_pose}
  \end{subfigure}
  \hfill
  \begin{subfigure}[b]{1.0\textwidth}
  \centering
    \includegraphics[trim=150 150 150 150,clip,width=\textwidth]{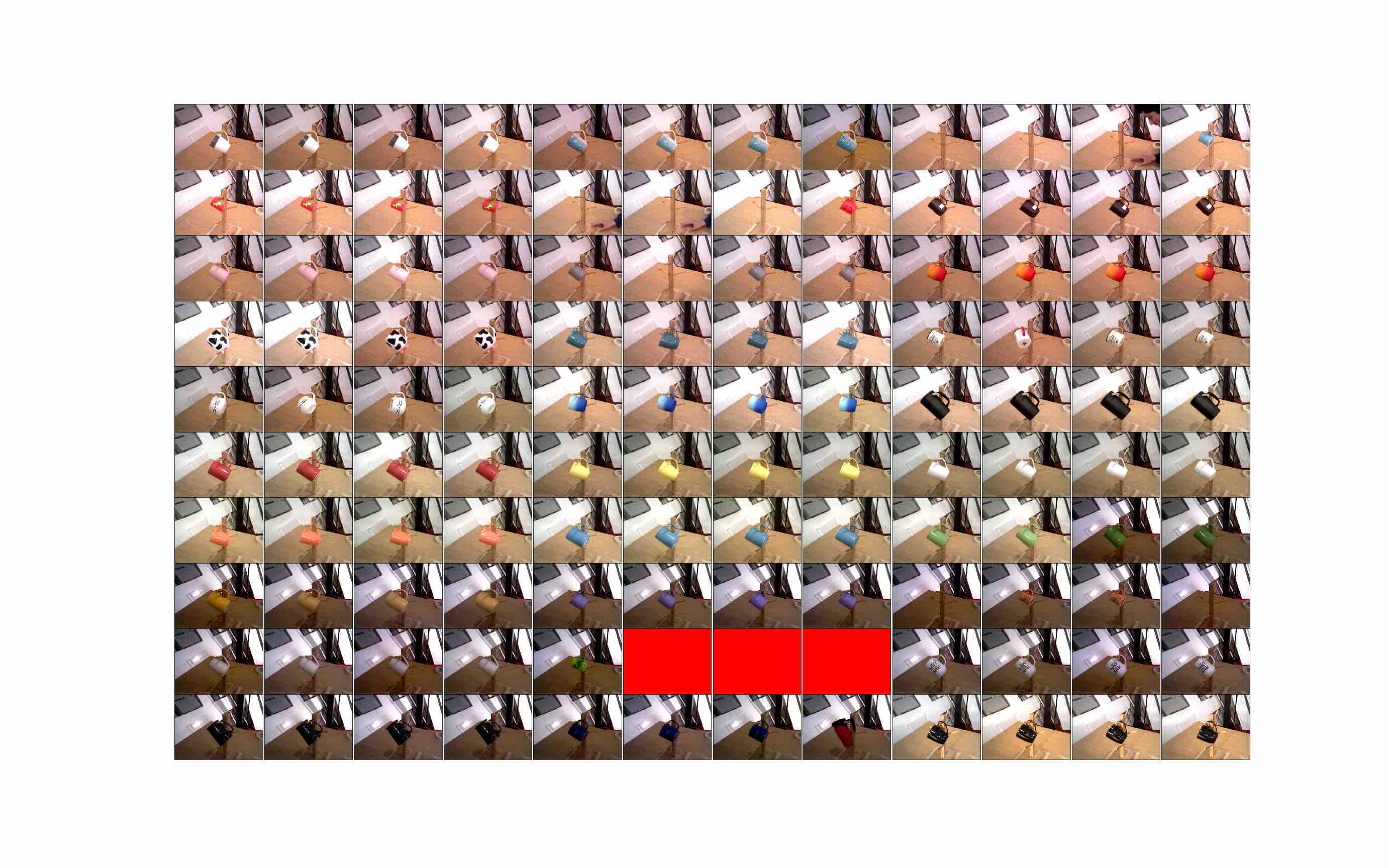}
    \caption{Image of the mug rack after running the kPAM pipeline. Red images indicate trials where no image of the final placement was captured due to an upstream failure of the pipeline causing the trial to be aborted.}
    \label{fig:mug_on_rack_tiled_end_pose}
  \end{subfigure}
  \caption{Before and after images of the mug on rack experiments for all 120 trials.}
  \label{fig:composite_images_mug_on_rack}
\end{figure}

\begin{figure}[!tbp]
\centering
  \begin{subfigure}[b]{1.0\textwidth}
  \centering
    \includegraphics[trim=150 150 150 150,clip,width=\textwidth]{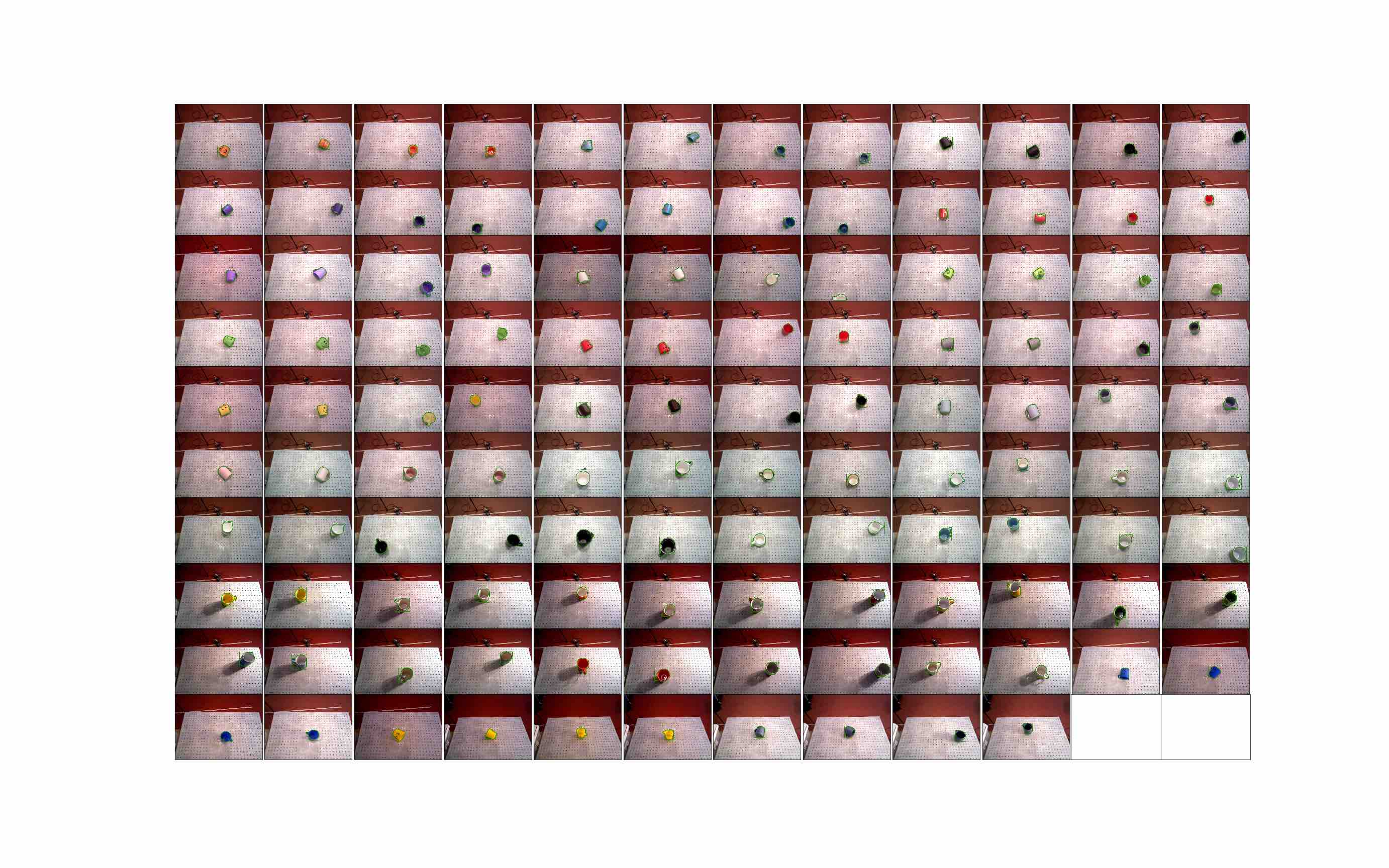}
    \caption{RGB images used for keypoint detection in mug on shelf experiments}
    \label{fig:mug_on_shelf_tiled_start_pose}
  \end{subfigure}
  \hfill
  \begin{subfigure}[b]{1.0\textwidth}
  \centering
    \includegraphics[trim=150 150 150 150,clip,width=\textwidth]{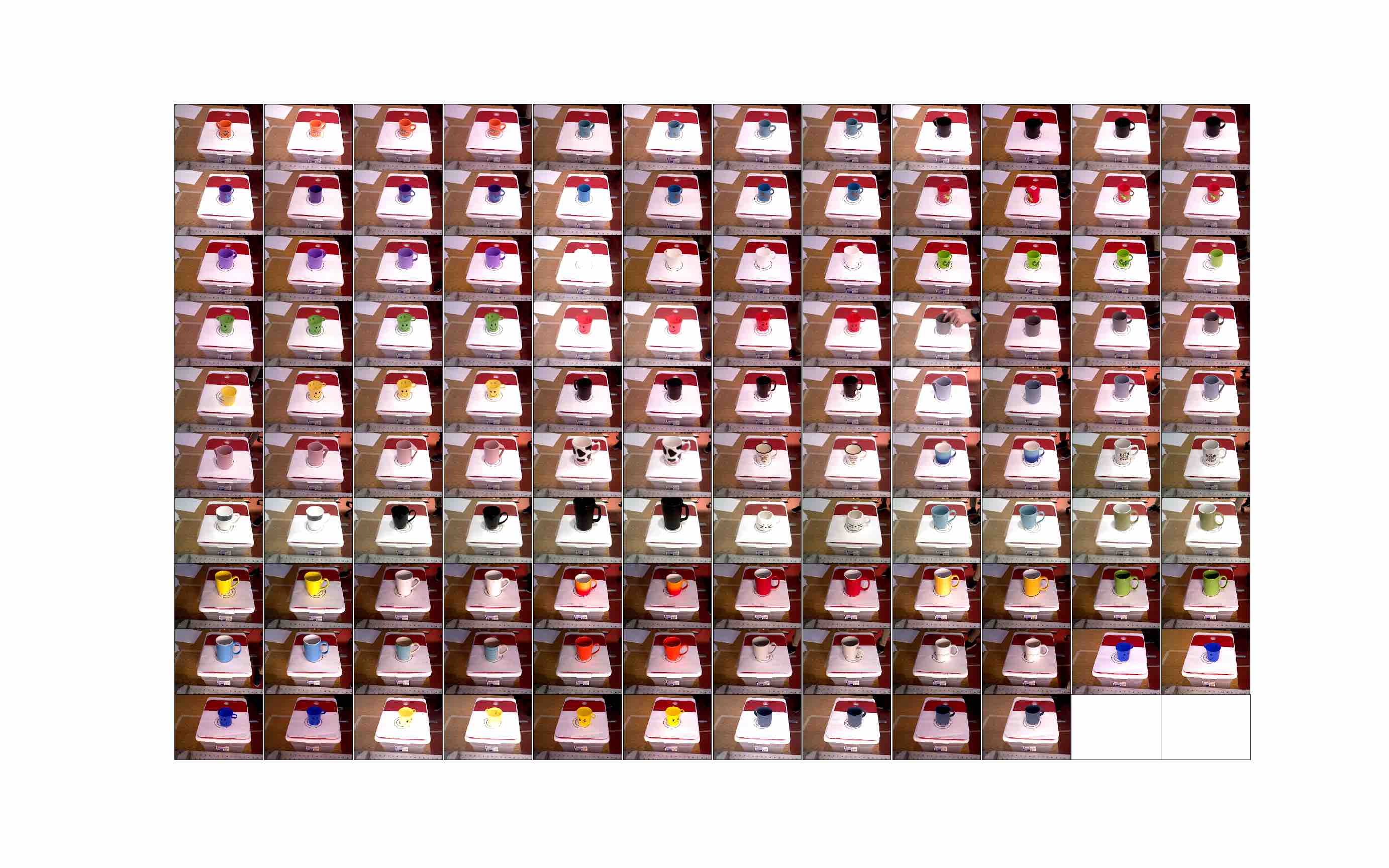}
    \caption{Image of the mug shelf after running the kPAM pipeline. Red images indicate trials where no image of the final placement was captured due to an upstream failure of the pipeline causing the trial to be aborted.}
    \label{fig:mug_on_shelf_tiled_end_pose}
  \end{subfigure}
  \caption{Before and after images of the mug on shelf experiments for all 118 trials.}
  \label{fig:composite_images_mug_on_shelf}
\end{figure}
Figures \ref{fig:composite_images_shoe_on_shelf}, \ref{fig:composite_images_mug_on_rack}, \ref{fig:composite_images_mug_on_shelf} illustrate the results of experiments. These figures containing tiled images showing thee initial RGB image used for keypoint detection, along with an image of the object after running the kPAM pipeline. In the following sections we discuss more details related to the mug on shelf and mug on rack experiments.

\subsection{Mugs Upright on Shelf}
Results for the mug on shelf experiment are detailed in Figure \ref{fig:quant_results}. A trial was classified as a sucess if the mug ended up upright on the shelf with it's bottom center keypoint within 5cm of the target location. Out of 118 trials we experienced 2 failures. One failure was due to a combination of inaccurate keypoint detections together with the mug being torqued as it was grasped. Since we only have a wrist mounted camera we cannot re-perceive the object to compensate for the fact that the object moves during the grasping process. As discussed in Section \ref{sec:limitations} this could be alleviated by adding an externally mounted camera.

\begin{figure}[t]
\centering
\includegraphics[width=0.6\textwidth]{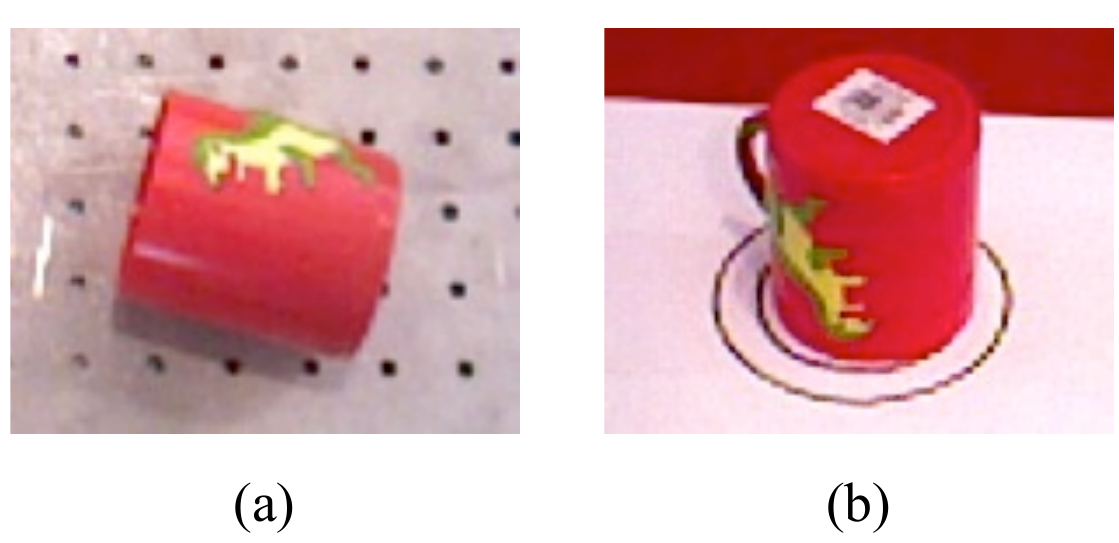}
\caption{\label{fig:mug_upside_down} (a) The RGB image for the single failure trial of the mug on shelf task that led to the mug being put in an incorrect orientation. In this case the keypoint detection confused the top and bottom of the mug and it was placed upside down. (b) The resulting upside down placement of the mug.}
\end{figure}

The other failure was resulted from the mug being placed upside down. Figure \ref{fig:mug_upside_down} shows the RGB image used for keypoint detection, along with the final position of the mug. As discussed in Section \ref{subsec:mugs_on_shelf} this failure occurred because the keypoint detection confused the top and bottom of the mug. Given that the image was taken from a side view where the handle is occluded and it is difficult to distinguish top from bottom is understandable that the keypoint detection failed in this case. There are several ways to deal with this type of issue in the future. One approach would be to additionally predict a confidence value for each keypoint detection. This would allow us to detect that we were uncertain about the keypoint detections in Figure \ref{fig:mug_upside_down} (a). We could then move the robot and collect another image that would allow us to unambiguously detect the keypoints.

\subsection{Hang mug on rack by its handle}

\begin{figure}[t]
\centering
\includegraphics[width=0.9\textwidth]{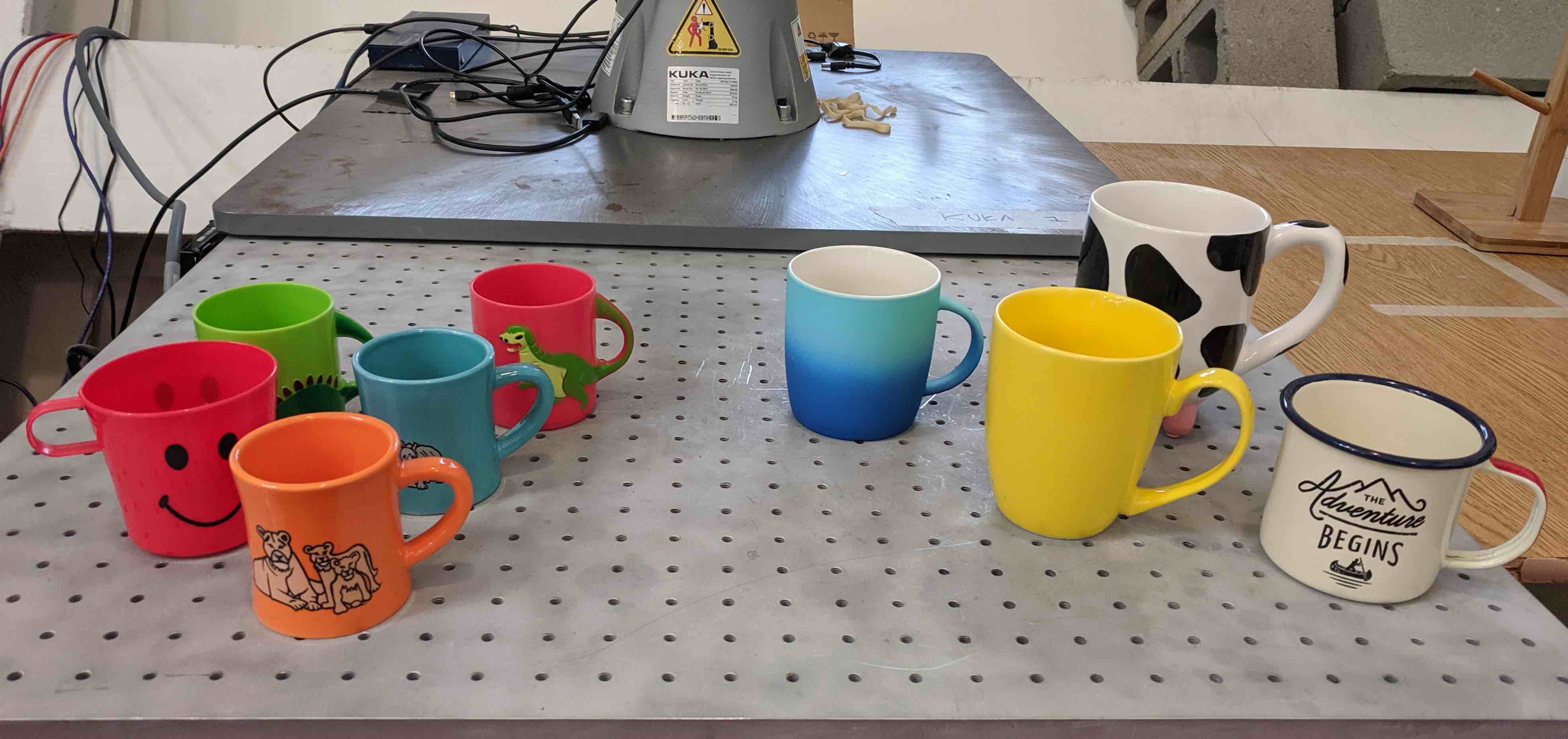}
\caption{\label{fig:mug_small_regular_comparison} The 5 mugs on the left are the test mugs used in experiment that were characterized as \emph{small}. For comparison the four mugs on the right are part of the \emph{regular} category.}
\end{figure}

As discussed in Section \ref{subsec:mug_rack_experiments} mugs were divided into two groups, \emph{regular} and \emph{small}, based on their size. A mug was characterized as \emph{small} if the handle had a minimum dimension (either height or width) of less than 2cm. Examples of mugs from each category are shown in Figure \ref{fig:mug_small_regular_comparison}. Mugs with such small handle sizes presented a challenge for our manipulation pipeline since hanging them on the rack requires increased precision.

% \newpage

% {\small
% %\bibliographystyle{styles/bibtex/spmpsci.bst}

% \bibliographystyleAppendix{plain}
% \bibliographyAppendix{paper.bib}
% }

\end{document}